\documentclass[11pt, a4paper,twoside,reqno]{amsart}

\usepackage{amsthm}
\usepackage[english]{babel}

\makeatletter
\def\@settitle{\begin{center}%
		\baselineskip14\p@\relax
		\normalfont\LARGE\scshape\bfseries
		\@title
	\end{center}%
}

\def\@setauthors{%
  \begingroup
  \def\thanks{\protect\thanks@warning}%
  \trivlist
  \centering\small \@topsep30\p@\relax
  \advance\@topsep by -\baselineskip
  \item\relax
  \author@andify\authors
  \def\\{\protect\linebreak}%
  \authors%
  \ifx\@empty\contribs
  \else
    ,\penalty-3 \space \@setcontribs
    \@closetoccontribs
  \fi
  \endtrivlist
  \endgroup
}

\makeatother

\makeatletter

\def\subsection{\@startsection{subsection}{2}%
	\z@{.5\linespacing\@plus.7\linespacing}{.5\linespacing}%
	{\normalfont\large\bfseries}}

\def\subsubsection{\@startsection{subsubsection}{3}%
	\z@{.5\linespacing\@plus.7\linespacing}{.5\linespacing}%
	{\normalfont\itshape}}

\usepackage[usenames, dvipsnames]{xcolor}
\definecolor{darkblue}{rgb}{0.0, 0.0, 0.45}
\usepackage[hidelinks]{hyperref}
\hypersetup{colorlinks	= true,
raiselinks	= false,
linkcolor	= darkblue, 
citecolor	= Mahogany,
urlcolor	= ForestGreen,
pdfauthor	= {Mohammad Boveiri},
pdftitle	= {},
pdfkeywords	= {},
pdfsubject	= {},
plainpages	= false}
\usepackage{dsfont,amssymb,amsmath, graphicx, multirow} 
\usepackage{amsfonts,dsfont,mathtools, mathrsfs,amsthm} 
\usepackage[amssymb, thickqspace]{SIunits}
\usepackage{algorithm,algorithmicx,fancyhdr}
\usepackage{makecell}

\allowdisplaybreaks
\date{\today}

\usepackage[english]{babel}
\usepackage{subcaption}
\usepackage{aligned-overset}
\usepackage{setspace}
\usepackage{todonotes}
\usepackage{enumerate}
\usepackage{doi}
\usepackage{algorithm}
\usepackage{algpseudocode}
\usepackage{tabularx}
\usepackage{amssymb,amsfonts}
\usepackage{dsfont}
\usepackage{lipsum}
\usepackage{colortbl}
\usepackage{array}
\usepackage{enumitem}
\usepackage{graphicx}
\usepackage{comment}
\usepackage{xcolor}
\usepackage{appendix}
\usepackage{enumitem}
\usepackage{csquotes}
\usepackage[giveninits=true,
  maxnames = 99,
  backref=false,
  backend=biber,
  style=alphabetic,
  sortlocale=de,
  natbib=true,
  doi=false,
  isbn=false,
  url=false]{biblatex}
\newtheorem{theorem}{Theorem}
\newtheorem{proposition}{Proposition}
\newtheorem{lemma}{Lemma}
\theoremstyle{definition} 
\newtheorem{remark}{Remark}

\newcommand{\T}{\mathcal{T}}

\title[Variance-Reduced Cascade Q-learning]{Variance-Reduced Cascade Q-learning: \\ Algorithms and Sample Complexity}
\author{Mohammad Boveiri and Peyman Mohajerin Esfahani}
\thanks{The authors are with the Delft Center for Systems and Control ({\tt\small \{M.Boveiri, P.MohajerinEsfahani\}@tudelft.nl}), Delft University of Technology, Delft, The Netherlands. This work is partially supported by the ERC grant TRUST-$949796$. }
\date{May 24, 2025}

\begin{document}

\begin{abstract}
We study the problem of estimating the optimal Q-function of $\gamma$-discounted Markov decision processes (MDPs) under the synchronous setting, where independent samples for all state-action pairs are drawn from a generative model at each iteration. We introduce and analyze a novel model-free algorithm called Variance-Reduced Cascade Q-learning (VRCQ). VRCQ comprises two key building blocks: (\romannumeral 1) the established direct variance reduction technique and (\romannumeral 2) our proposed variance reduction scheme, Cascade Q-learning. By leveraging these techniques, VRCQ provides superior guarantees in the $\ell_\infty$-norm compared with the existing model-free stochastic approximation-type algorithms. Specifically, we demonstrate that VRCQ is minimax optimal. Additionally, when the action set is a singleton (so that the Q-learning problem reduces to policy evaluation), it achieves non-asymptotic instance optimality while requiring the minimum number of samples theoretically possible. Our theoretical results and their practical implications are supported by numerical \mbox{experiments}.
\end{abstract}
\maketitle

\section{Introduction}
Markov decision processes (MDPs) and reinforcement learning (RL) are widely used mathematical frameworks for decision-making in dynamic and uncertain environments, with an extensive history of research~\cite{Bellman, Watkins, RL-Book-Bertsekas.2, RMDP, F.Natural,Tac.1997}. In recent years, thanks to the surge in available data and computing power, RL techniques have enjoyed tremendous success across a wide range of applications~\cite{Nature1,Nature2,S.levine,Nature3,Autonomous.Driving,Nature4,tac.12}.

Generally, there are two main approaches to RL: model-based~\cite [see, e.g.,][] {Minimax.bound,Kakade.model.based,Li.model_base,OR3} and model-free~\cite [see, e.g.,][]{Bubek,Speedy,SA.Wainwright,VRQ.Wainwright,MDVI,Q-learning-PL,TAC.free,modelfree.2}. In model-based RL, we first learn a model of the MDP using a batch of state-transition samples and then use this model to form a control policy. On the other hand, model-free approaches directly update either the value function, representing the expected reward starting from each state, or the policy, which is the mapping from states to their actions. Given that model-free RL algorithms operate online, demand less storage space, and are more expressive, the majority of state-of-the-art RL developments have been within the model-free paradigm~\cite{Mnih, Mnih2, Trust.Region, Nature3}.

This paper focuses on model-free algorithms for finite state-action RL problems when we have access to a generative model of MDP~\cite{Kakade, Kearns}. That is, a sampling model or a simulator that produces independent samples for all state-action pairs. Our analysis is specifically focused on infinite-horizon discounted MDPs, with the states space $\mathcal{X}$,  the actions space  $\mathcal{U}$, and the discount factor $\gamma \in (0, 1)$. It is worth noting that although the generative setting is generally simpler than online RL settings~\cite{Tor,Bubek,Li.Asynchronous}, the results and techniques developed within this framework often extend to more complex settings~\cite[see, e.g.,][]{Minimax.bound,azar.regret}.
With the generative setting in mind, we introduce a novel model-free algorithm named Variance-Reduced Cascade Q-learning (VRCQ, Algorithm \ref{algorithm.2}) and analyze its $\ell_\infty$-based sample complexity, namely the number of samples required for VRCQ to yield an entrywise $\epsilon$-accurate estimate of the optimal Q-function.

The VRCQ algorithm consists of two building blocks: (\romannumeral 1) a direct variance reduction technique inspired by~\cite{VRQ.Wainwright,SVRG,VR.LeRoux}, and (\romannumeral 2) our novel variance reduction scheme, Cascade Q-learning (Algorithm \ref{algorithm.1}). Thanks to these methods, VRCQ offers better theoretical guarantees in the $\ell_\infty$ norm compared with existing model-free algorithms. To demonstrate this, we examine the sample complexity of VRCQ from two perspectives: (\romannumeral 1) the minimax viewpoint, which provides bounds that hold uniformly over large classes of models~\cite{Minimax.bound,Speedy, VRQ.Wainwright, Q-learning-tight, MDVI}, and (\romannumeral 2)  the instance-dependent perspective~\cite{Khamaru.TD,Khamaru.QL,Q-learning-PL, Pananjady, pmlr.instance,instance.mathor}, offering bounds that hold locally around each problem instance. In the latter case, to simplify the analysis and facilitate comparison with recent works~\cite{Khamaru.TD,Khamaru.QL, Root.SA, Q-learning-PL}, we focus on the scenario where the action set $\mathcal{U}$ is a singleton (i.e., $|\mathcal{U}|=1$).

\subsection*{Main Contributions and Connection to Prior Work}
The following outlines our main contributions and positions them within the relevant literature.

(\romannumeral 1 ) \textbf {Cascade Q-learning (CQ) and its direct variance-reduced extension.} We introduce a novel variance reduction scheme named Cascade Q-learning (CQ, Algorithm \ref{algorithm.1}). By establishing a non-asymptotic upper bound on the performance of CQ, we demonstrate that, thanks to its unique structure, it mitigates the impact of noise (Proposition \ref{proposition.1}). Moreover, employing the CQ scheme and the direct variance-reduced technique~\cite{VRQ.Wainwright,wang.Value,SVRG}, we propose a novel model-free RL algorithm called Variance-Reduced Cascade Q-learning (VRCQ, Algorithm \ref{algorithm.2}). VRCQ follows an epoch-based structure akin to standard variance reduction schemes, such as SVRG~\cite{SVRG} and the variance-reduced Q-learning~\cite{VRQ.Wainwright}. In each epoch, we run the CQ algorithm, but we recenter our updates to reduce their variance.

(\romannumeral 2 ) \textbf {Geometric convergence rate over epochs with shorter epoch lengths.} Similar to standard direct variance reduction schemes, VRCQ exhibits a geometric convergence rate as a function of the epoch number (Theorem \ref{theorem.1}). VRCQ achieves this while utilizing only the order of $(1-\gamma)^{-2}$ samples (up to a logarithmic factor) within the inner loop of each epoch (the epoch length). In contrast, the variance-reduced Q-learning proposed in~\cite{VRQ.Wainwright} requires the order of $(1-\gamma)^{-3}$ samples (up to a logarithmic factor) as the epoch length to achieve a similar outcome. (see Theorem \ref{theorem.1} and Remark \ref{Remark.2} for more details).

  (\romannumeral 3) \textbf {Minimax optimality.} We  consider the class $\mathcal{M}(\gamma,r_{\text{max}})$ of optimal Q-functions that can be obtained from $\gamma$-discounted MDPs with a reward function  bounded by $r_{\text{max}}$. Over this class, we show that VRCQ requires at most $\mathcal{O}( \frac{r^2_{\text{max}} \log\big(\frac{D}{\delta} \log (\frac{r_{\text{max}}}{(1-\gamma)\epsilon}) \big) } {\epsilon^2(1-\gamma)^3})$ samples to return an $\epsilon$-accurate solution with probability at least $1-\delta$, where $ \epsilon \in (0, 1]$ and $D= |\mathcal{X}|\times |\mathcal{U}|$ . This upper bound matches the minimax lower bound  $ \Omega\big(\frac{r^2_{\text{max}} \log\big(\frac{D}{\delta} \big) } {\epsilon^2(1-\gamma)^3}\big)$~\cite{Minimax.bound} up to a $\log\Big(\log(\frac{r_{\text{max}}}{(1-\gamma)\epsilon})\Big)$ factor. In contrast, the worst case sample complexity of Q-learning~\cite{Q-learning-tight} and Speedy Q-learning~\cite{Speedy} scale as $(1-\gamma)^{-4}$. Mirror descent value iteration~\cite{MDVI} has the worst-case optimal cubic dependency on $(1-\gamma)^{-1}$ when $\epsilon \in (0, \sqrt{1-\gamma}\,]$. This implies that $\epsilon$ must be sufficiently small when $\gamma$ is close to $1$. Variance-reduced Q-learning~\cite{VRQ.Wainwright} has the worst-case sample complexity  $\mathcal{O}( \frac{r^2_{\text{max}} \log(\frac{1}{1-\gamma}) \log\big(\frac{D}{\delta} \log (\frac{r_{\text{max}}}{(1-\gamma)\epsilon}) \big) } {\epsilon^2(1-\gamma)^3})$ for  $ \epsilon \in (0, 1]$. As we can see, in the worst-case scenario, VRCQ outperforms variance-reduced Q-learning by a logarithmic factor in the discount complexity $(1-\gamma)^{-1}$. This improvement is the direct consequence of the previously mentioned feature of the VRCQ algorithm, namely achieving geometric convergence over epochs with shorter epoch lengths. While this feature only results in a logarithmic improvement in the worst-case setting, we will demonstrate that by considering a stronger criterion for optimality, known as instance optimality~\cite{Khamaru.TD,Khamaru.QL,instance.mathor,Pananjady}, the distinction in performance between these two algorithms becomes more apparent. A detailed comparison between VRCQ and other algorithms from the minimax perspective is provided in~Remark~\ref{Remark.3}.

(\romannumeral 4) \textbf{Non-asymptotic instance optimality.} We study the instance-dependent behavior of VRCQ when the action set consists of a single element ($|\mathcal{U}|=1$), so that the problem of estimating the optimal Q-function reduces to the policy evaluation problem. In this setting, we provide an instance-dependent upper bound on the $\ell_\infty$-error in the non-asymptotic regime and show that the VRCQ algorithm is instant optimal when the number of samples scales as $(1-\gamma)^{-2}$ (Theorem \ref{theorem.3}). This sample size requirement matches the lower bound developed in~\cite{Khamaru.TD}. By comparison, Polyak-Ruppert averaged Q-learning with both polynomial step size and rescaled linear step size is suboptimal in the non-asymptotic setting~\cite{Khamaru.TD, Q-learning-tight}. Additionally, variance-reduced Q-learning~\cite{VRQ.Wainwright} is instance optimal when the number of samples scales as $(1-\gamma)^{-3}$~\cite{Khamaru.TD}. Remark \ref{remark.5} provides a more detailed comparison between VRCQ and other algorithms from the instance-dependent viewpoint.

\scriptsize
\begin{table}[h!]
\centering
\begin{tabularx}{ 1 \textwidth}
 { 
  | >{\centering\arraybackslash} X 
  | >{\centering\arraybackslash} X 
  | >{\centering\arraybackslash} X |}
  \hline
 {Algorithm} &  worst-case sample complexity  & $\epsilon$-range  \\
 \hline
  Q-learning\cite{Q-learning-tight}   & $ {\epsilon^{-2}(1-\gamma)^{-4} }$  &    $\epsilon \in (0,  1] $  \\
 \hline
 Speedy Q-learning~\cite{Speedy} &   $ {\epsilon^{-2}(1-\gamma)^{-4} }$   &   $\epsilon \in (0,  1] $   \\
 \hline
  Mirror descent value iteration~\cite{MDVI}  &    $ \epsilon^{-2}(1-\gamma)^{-3}$   &   $\epsilon \in  (0, \sqrt{1-\gamma}\,] $ \\  
 \hline 
  {VRCQ} (this paper)     &   ${\epsilon^{-2}(1-\gamma)^{-3} }$ 
   &  $\epsilon \in (0,  1] $  \\
 \hline
\end{tabularx}
\caption{Upper bounds on the worst-case sample complexity for some algorithms. All logarithmic factors are omitted in the table to simplify the expressions. }
\label{table:1}
\end{table}
\normalsize

\begin{table}[h!]
\centering
\small
\begin{tabularx}{ 1 \textwidth}
 { 
  | >{\centering\arraybackslash}X 
  | >{\centering\arraybackslash}X 
  | >{\centering\arraybackslash}X
  | >{\centering\arraybackslash}X | }
\hline
  {Algorithm} & Upper bound on the $\ell_\infty$-error &  Sample size requirement   \\
\hline
  PR averaged Q-learning with rescaled linear step~size \mbox{\cite{Khamaru.TD,Q-learning-PL}} & \qquad \qquad \qquad \qquad \qquad \qquad\qquad \qquad \qquad \qquad \qquad Suboptimal & suboptimal even in the asymptotic regime \mbox{\cite[Sec. 5]{Q-learning-PL}}\\
\hline
  PR averaged Q-learning with polynomial step~size \mbox{\cite{Khamaru.TD,Q-learning-PL}}   &$\sqrt{\log(D)}\Big(\gamma v (\mathcal{P})+\rho(\mathcal{P})\Big)$, optimal up to a logarithmic factor in the dimension  & Asymptotically optimal, but suboptimal in the non-asymptotic regime \\ 
 \hline 
  Variance-reduced Q-learning \cite{VRQ.Wainwright,Khamaru.TD,Khamaru.QL}  & $\sqrt{\log(D)}\Big(\gamma v (\mathcal{P})+\rho(\mathcal{P})\Big)$, optimal up to a logarithmic factor in the dimension    &  \qquad \qquad \qquad \qquad \qquad \qquad\qquad \qquad \qquad \qquad \qquad $\mathcal{O}\big(\frac{\log(D) }{(1-\gamma)^3}\big)$, suboptimal\\
 \hline
 \qquad \qquad \qquad \qquad \qquad \qquad \qquad VRCQ (this paper)   &  $\sqrt{\log(D)}\Big(\gamma v (\mathcal{P})+\rho(\mathcal{P})\Big)$, optimal up to a logarithmic factor in dimension  &   $\mathcal{O}\big(  \frac{\log(D)}{(1-\gamma)^2} \big)$, optimal up to a logarithmic factor in the dimension  \\
 \hline 
    Lower bound~\cite{Khamaru.TD}   &   $\gamma v (\mathcal{P})+\rho(\mathcal{P})$   &  $ \Omega \big( \frac{1}{(1-\gamma)^2} \big)$   \\
 \hline
\end{tabularx}\\
\caption{Instant-dependent bounds on the $\ell_\infty$-norm of error in the non-asymptotic regime when $|\mathcal{U}|=1$.}
\label{table:2}
\end{table}
\normalsize

The remainder of this paper is organized as follows. In Section \ref{sec:background}, we present a review of fundamental concepts related to MDPs and RL. In Section \ref{sec:main result}, which serves as the central part of this paper, we begin by introducing a novel scheme called cascade Q-learning (CQ), designed to mitigate the impact of noise throughout the horizon. Subsequently, we present a direct variance-reduced extension of CQ, referred to as VRCQ, and examine its sample complexity from both the minimax and instance-dependent perspectives. Section \ref{sec:simulation} includes illustrative examples and simulation results. In Section \ref{sec: Conclusion}, we discuss some conclusions and outline potential future directions. The proofs of the main results are provided in Section \ref{sec: proofs}.

\textbf{Notations.} Throughout the paper, we use the following notations. The symbols $\mathbb{R}$ and $\mathbb{R}_{>0}$ represent the sets of real numbers and positive real numbers, respectively. The cardinality of a set $\mathcal{S}$ is denoted by $|\mathcal{S}|$. Given matrices $A, B \in \mathbb{R}^{n \times m}$, $A \leq B$ means $A_{ij} \leq B_{ij}$ for all $i=1,..,n$ and $j=1,..,m$. The $\ell_\infty$-norm and span seminorm of $A$ are, respectively, defined as $\|A\|_{\infty} := \underset{i,j}{\max }\; |A_{ij}|$ and $\|A\|_{\text{span}}:=\underset{i,j}{\max}\; A_{ij}-\underset{i,j}{\min}\; A_{ij}$. We use $\mathds{1}$ to denote the all-ones matrix. The function \mbox{$\log: \mathbb{R}_{>0} \rightarrow \mathbb{R}$} represents the natural logarithm.

\section{Setting and Problem Description} 
\label{sec:background}
This section briefly reviews some standard concepts of MDPs and RL. For further background on these topics, we refer the reader to several books (e.g.,~\cite{RL-Book_Bertsekas, RL_Book_Puterman, Szepesvari, RL-Book_Sutton}). A discounted MDP is a quintuple $(\mathcal{X}, \mathcal{U}, \mathbb{P}, r, \gamma )$, where $\mathcal{X}$ is a finite set of possible states, $\mathcal{U}$ is a finite set of possible actions, $\mathbb{P}=\{\mathbb{P}_u(\cdot\,|\,x) \;|\; (x,u) \in \mathcal{X} \times \mathcal{U}\}$ is the collection of state-action probability transition functions (when in state $x$, executing an action $u$ causes a transition to the next state drawn randomly from the transition function $\mathbb{P}_u(\cdot\,|\,x)$), $r : \mathcal{S} \times \mathcal{U} \rightarrow \mathbb{R} $ is the reward function (i.e., $r(x,u)  $ is the immediate reward collected in state $x \in \mathcal{X}$ when action $u \in \mathcal{U}$ is taken.), and $\gamma\in (0,1)$ indicates the discount factor. A deterministic policy $\pi: \mathcal{X} \rightarrow \mathcal{U}$ is a map from the set of states $\mathcal{X}$ to the set of actions $\mathcal{U}$. The action-value function or Q-function of a given policy $\pi$ is defined as
\begin{equation*}
\Theta^\pi(x,u) := \mathbb{E} \left[\sum_{n=0}^{\infty}\gamma^n r(x_n,u_n) \;|\; (x_0,u_0)=(x,u)\right],
\end{equation*}
where $u_n=\pi(x_n)$ for all $n\geq 1$, and the expectation is evaluated with respect to the randomness of the MDP trajectory. Moreover, $\Theta^\star (x,u) =\underset{\pi}{\sup} \;\Theta^\pi (x,u) $ and $\pi^\star(x)=\arg\underset{u}{\max}\; \Theta^\star(x,u)$ are called the optimal Q-function and optimal policy, respectively.
It is well known from the theory of MDPs that $\Theta^\star$ is the unique fixed point of the Bellman operator. The Bellman operator is mapping from $\mathbb{R}^{|\mathcal{X}|\times|\mathcal{U}|}$ to itself, whose $(x,u)-$entry is given by
\begin{equation*}
    \mathcal{T}(\Theta)(x,u):=r(x,u)+\gamma \mathbb{E}_{\Bar{x}} \max_{\Bar{u}} \Theta(\Bar{x},\Bar{u}),\ \ \ \  \ \text{where}\ \ \ \Bar{x} \sim \mathbb{P}_u(\cdot\,|\,x).
\end{equation*}

In the context of RL, the probability transition functions $\mathbb{P}=\{ \mathbb{P}_u(\cdot\,|\,x) \;|\; (x,u) \in (\mathcal{X},\mathcal{U})\}$ are unknown. As a result, the Bellman operator cannot be exactly evaluated. In the generative setting of RL, at each iteration $n$, we observe a sample $x_n(x,u)$ drawn from the transition function $\mathbb{P}_u(\cdot\,|\,x)$ for every pair $(x,u)$. Moreover, we assume at each iteration we have access to a noisy observation $\hat{r}_n(x,u)$ of the reward function with mean $r(x,u)$, and $\sigma_r$-sub-Gaussian tails. The rewards $\{\hat{r}(x,u)\}_{(x,u) \in (\mathcal{X},\mathcal{U})}$ are independent across all state-action pairs, as well as the randomness in $x_n(x,u)$. The objective is to compute an approximation of the optimal Q-function based on these observations. The empirical Bellman operator at iteration $n$ is defined as 
\begin{equation*}
    \widehat{\T}_n(\Theta)(x,u):=\hat{r}_n(x,u)+ \gamma \max_{\Bar{u}} \Theta(x_n,\Bar{u}),\ \ \ \ \ \text{where} \ \ \ x_n\equiv x_n(x,u) \sim \mathbb{P}_u(\cdot\,|\,x).
\end{equation*}
Note that the empirical Bellman operator is an unbiased estimation of the Bellman operator, i.e., $\mathbb{E}\widehat{\T}_n (\Theta)=\mathcal{T}(\Theta)$. Moreover, the Bellman and empirical Bellman operators are $\gamma$-contractive in the $\ell_\infty$-norm. The matrix $W_n := \widehat{\T}_n (\Theta^\star) - \mathcal{T}(\Theta^\star)$ denotes the effective noise or the Bellman noise associated with the operator $\widehat{\T}_n$. It indicates the failure of $\widehat{\T}_n$ to maintain $\Theta^\star$ as its fixed point~\cite[see][Sec. 2.2]{SA.Wainwright}. It is worth noting that the $(x,u)$-entry of $W_n$ can be written as
\begin{equation*}
    W_n (x,u)= \Big(\hat{r}_n (x,u) - r(x,u) \Big) + \gamma \Big( \max_{\Bar{u}} \Theta^\star(x_n,\Bar{u}) -  \mathbb{E}_{\Bar{x}} \max_{\Bar{u}} \Theta^\star(\Bar{x},\Bar{u})   \Big).
\end{equation*}
The first term on the right-hand side of $W_n(x,u)$ is a sub-gaussian random variable with variance at most $\sigma_r$, and the second term is bounded in absolute value by $\gamma \|\Theta^\star\|_{\text{span}}$ and has the variance
\begin{equation*}
    \sigma (\Theta^\star) (x,u) := \gamma^2 \mathbb{E}_{\Tilde{x}} \Big( \max_{\Bar{u}} \Theta^\star(\Tilde{x},\Bar{u})- \mathbb{E}_{\Bar{x}} \max_{\Bar{u}} \Theta^\star(\Bar{x},\Bar{u}) \Big)^2.
\end{equation*}
Here the expectations $\mathbb{E}_{\Bar{x}}$ and $\mathbb{E}_{\Tilde{x}}$ are both computed over $\mathbb{P}_u(\cdot\,|\,x)$. The matrix of variances $\sigma(\Theta^\star)$ is referred to as the effective variance matrix, and it plays a central role in the non-asymptotic analysis of stochastic approximation-type RL algorithms~\cite{SA.Wainwright,VRQ.Wainwright,Q-learning-tight,Q-learning-PL}.

As the final preliminary point discussed in this section, we note that in the generative setting, since each iteration involves drawing $D=|\mathcal{X}|\times |\mathcal{U}|$ samples, the number of samples is a factor of $D$ larger than the number of iterations.

\section{Main Results} \label{sec:main result}

In this section, we first introduce a novel scheme called Cascade Q-learning (Algorithm \ref{algorithm.1}) and motivate it from a variance reduction standpoint. We then integrate this scheme with the direct variance-reduced technique~\cite{VR.LeRoux, SVRG, Katyusha, wang.Value, VRQ.Wainwright} to develop an algorithm with superior theoretical guarantees (Algorithm \ref{algorithm.2}).

\subsection{Cascade Q-Learning: A New Scheme to Reduce the Effect of Noise}

The pseudo-code of the Cascade Q-learning (CQ) is shown in Algorithm \ref{algorithm.1}. At each iteration, CQ evaluates the empirical Bellman operator $\widehat{\T}_n$ at the point $Y_{n+1}=(1-\lambda)Y_n + \lambda Z_n$. It then calculate $Z_{n+1}$ based on the update rule $Z_{n+1}=(1-\lambda)Z_n + \lambda \widehat{\T}_n (Y_{n+1})$, and average the iterates $\{Y_{i+1}\}_{i=1}^{n}$. To gain a better understanding of the algorithm, it is noteworthy that if we replace the update rule $Y_{n+1}=(1-\lambda) Y_n + \lambda Z_n$ with $Y_{n+1}=Z_n$, then CQ transforms into Polyak-Ruppert (PR) averaged ~\cite{PL.Polak} Q-learning with the constant step size $\lambda$.
Therefore, generally speaking, CQ can be seen as PR averaged Q-learning, coupled with an additional filtering step (or a momentum term), given by $Y_{n+1}=(1-\lambda) Y_n + \lambda Z_n$. As we demonstrate shortly, thanks to its underlying structure, CQ outperforms Q-learning in its ability to handle noise fluctuations. The following proposition provides an upper bound on the performance of the CQ algorithm. \par

\begin{algorithm}
\caption{Cascade Q-learning (CQ)}
\label{algorithm.1}
\begin{algorithmic}
\Require $N_e$ and $\Theta_0$
\State $Y_1=Z_1= \Theta_0$
\For{$n=1,....,N_e$}
\State $\quad Y_{n+1}=(1-\lambda) Y_{n} + \lambda Z_{n}$
\State $\quad Z_{n+1}=(1-\lambda) Z_{n} + \lambda \widehat{\T}_n (Y_{n+1})$
\State $\quad \;\;\;\;\Theta_{n}=\frac{1}{n}\sum_{i=1}^n Y_{i+1}$
\EndFor
\end{algorithmic}
\end{algorithm}

\begin{proposition}[Non-asymptotic guarantee for Cascade Q-learning]
\label{proposition.1}
 Consider an MDP with discount factor $\gamma$ and optimal Q-function $\Theta^\star$. Suppose we run Algorithm \ref{algorithm.1} from the initialization $\Theta_0$ for $N_e$ iterations with the constant step size $\lambda =\frac{1}{\sqrt{N_e}}$. Then, we have
\begin{align}\label{Eq:Theorem1.a}
 \mathbb{E}\|\Theta_{N_e}-\Theta^\star\|_{\infty} \leq \frac{2\|\Theta_0-\Theta^\star\|_{\infty}}{ (1-\gamma)\sqrt{N_e}}+\frac{2 \gamma\log(2D)\|\Theta^\star\|_{\text{span}}}{3 (1-\gamma)N_e}+\frac{2 \sqrt{2\log(2D)}\big(\|\sigma(\Theta^\star)\|_{\infty} +\sigma_r \big)}{(1-\gamma)\sqrt{N_e}}.
\end{align}

\end{proposition}

Some comments on Proposition \ref{proposition.1} are in order. The first term on the right-hand side of the above inequality indicates the initialization error, while the second and third terms originate from the fluctuations of the Bellman noise $W_n = \widehat{\T}_n(\Theta^\star) - \T(\Theta^\star)$ in the algorithm. Furthermore, it follows directly from the above inequality that by running CQ for 
\begin{equation}\label{Eq:Theorem1.c}
  N_e \geq  c \Big( \frac{\|\Theta_0-\Theta^\star \|_{\infty}^2 }  {(1-\gamma)^2\epsilon^2}+\frac{\gamma\log(D) \|\Theta^\star\|_{\text{span}} }{(1-\gamma)\epsilon}+\frac{\log(D)  \big(\|\sigma(\Theta^\star)\|_{\infty}^2 +\sigma_r^2\big) }  {(1-\gamma)^2\epsilon^2}\Big)  
\end{equation}
iterations, we have $\mathbb{E}\|\Theta_{N_e}-\Theta^\star\|\leq \epsilon$, where c is a universal constant.

\textbf{Cascade Q-learning versus standard Q-learning.} Consider the Q-learning algorithm given by $\Theta_{n+1}=(1-\lambda_n)\Theta_{n}+\lambda \widehat{\T}_n (\Theta_n)$, with the rescaled linear step size $\lambda_n=\frac{1}{1+(1-\gamma)n}$. Moreover, assume $\sigma_r=0$. In~\cite{SA.Wainwright}, it is shown that running Q-learning for 
\begin{equation}\label{Eq:Comparison.Q-learning}
 N_e \geq c \Big(\frac{\|\Theta_0-\Theta^\star\|_{\infty}}{(1-\gamma) \epsilon} + \frac{\log(D) \|\Theta\|_{\text{span}}} {(1-\gamma)^2 \epsilon }+ \frac{\log^2(D) \big( \|\sigma(\Theta^\star)\|_{\infty}^2  \big)}{(1-\gamma)^3 \epsilon^2 }   \Big)
\end{equation}
iterations results in an $\epsilon$-accurate solution in expectation. Furthermore,~\cite{SA.Wainwright} provides an example demonstrating that \eqref{Eq:Comparison.Q-learning} is sharp, indicating that this bound is generally unimprovable. It is also worth mentioning that the last term on the right-hand side of \eqref{Eq:Comparison.Q-learning}, representing the variance of the Bellman noise, is the dominant term. A close examination of \eqref{Eq:Theorem1.c} and \eqref{Eq:Comparison.Q-learning} reveals that the dependency on the horizon, $(1-\gamma)^{-1}$, in the last two terms (noise-related terms) in CQ, as compared to Q-learning, has been improved. Therefore, in general terms, CQ can be considered a form of variance reduction, where the impact of noise through the horizon $(1-\gamma)^{-1}$ has been reduced. Notably, this improvement has resulted in reducing the predominant dependence on the horizon from $(1-\gamma)^{-3}$ to $(1-\gamma)^{-2}$.

\subsection{Variance-Reduced Cascade Q-learning (VRCQ)}\label{VRCQ}

The full potential of cascade Q-learning becomes apparent when combined with the direct variance reduction technique~\cite{VR.LeRoux, SVRG, Katyusha, wang.Value, VRQ.Wainwright}. This method allows us to select step sizes that are independent of the total iteration number, resulting in a faster algorithm. 

In this section, we present a direct variance-reduced extension of CQ, referred to as VRCQ (Algorithm \ref{algorithm.2}). VRCQ follows an epoch-based structure akin to standard direct variance reduction schemes~\cite{SVRG,Katyusha,VRQ.Wainwright}. In each epoch, we run the CQ scheme but we recenter our updates to reduce their variance. Similar to variance-reduced Q-learning~\cite{VRQ.Wainwright}, this recentering employs an empirical approximation $\widetilde{\T}$ to the population Bellman operator $\mathcal{T}$. The complete description of VRCQ is summarized in Algorithm \ref{algorithm.2}. The overall algorithm is characterized by four key choices: the total number of epochs denoted as $M$; the sequence of step sizes $\{\lambda(m)\}_{m=0}^{M-1}$; the sequence of epoch lengths $\{N_e(m)\}_{m=0}^{M-1}$, and the sequence of recentering samples $\{N_\T(m)\}_{m=0}^{M-1}$. Our initial finding indicates that through a suitable choice of these parameters, we achieve a geometric convergence rate as a function of the epoch number. \par

\begin{algorithm}
\caption{Variance-Reduced Cascade Q-learning (VRCQ)}
\label{algorithm.2}
\begin{algorithmic}
 \For {$m=0,..., M-1$}
 \State  $\widetilde{\T}(\Theta_{m}) = \frac{1} {N_{\T}} \sum_{i=1}^{N_{\T}} \widehat{\T}_i (\Theta_m)$
  \State $Y_1=Z_1=\Theta_{m}$
 \For {$n=1,....,N_e$}
   \State $\quad Y_{n+1}  =(1-\lambda) Y_{n} + \lambda Z_{n}$
    \State $\quad Z_{n+1}  =(1-\lambda) Z_{n} + \lambda \left( \widehat{\T}_n (Y_{n+1})-\widehat{\T}_n(\Theta_m)+ \widetilde{\T} (\Theta_m)\right)$ 
    \EndFor
    \State $\Theta_{m+1}=\frac{1}{N_e} \sum_{n=1}^{N_e} Y_{n+1}$

\EndFor
\end{algorithmic}
\end{algorithm}

\begin{theorem}[Geometric convergence over epochs]
\label{theorem.1}
Consider an MDP with the discount factor $\gamma$ and optimal Q-function $\Theta^\star$. Let $\phi \in (0,1)$ be the desired convergence rate over epochs. suppose we run Algorithm \ref{algorithm.2} from initial point $\Theta_0=0$ for $M$ epochs (for $m=0,..., M-1$)
\begin{enumerate}[label=(\alph*)]
    \item \textbf{Convergence in expectation:} If we choose  $\lambda(m)=\frac{1}{\sqrt{N_e(m)}}$ , $ N_{\T}(m) \geq \frac{32\log(2D)}{\phi^{2m+2}(1-\gamma)^2}$, and $N_{e} (m)\geq \frac{13^2\log(2D)}{\phi^2(2-\phi^m)^2(1-\gamma)^2}$, then we have $\mathbb{E}\|\Theta_M-\Theta^\star \|_{\infty} \leq \phi^M \big(\|\Theta^\star\|_{\infty}+\sigma_r \big)$.\par  
    \item \textbf{Convergence with high probability:} By setting $N_{\T}(m) \geq \frac{32\log(\frac{10 MD}{\delta})}{\phi^{2m+2}(1-\gamma)^2}$, and $N_{e}(m)\geq \frac{338 \log(\frac{1690 MD}{\phi^2(2-\phi^m)^2(1-\gamma)^2\delta})}{\phi^2(2-\phi^m)^2(1-\gamma)^2}$, $\|\Theta_M-\Theta^\star \|_{\infty} \leq \phi^M \big(\|\Theta^\star\|_{\infty}+\sigma_r\big)$ holds with probability at least $1-\delta$.\par 
\end{enumerate}\par
\end{theorem}
\begin{remark}[Behavior of the parameters over epochs]\label{Remark.1}
As the epoch number $m$ increases, both the recentring sample size $N_{\T}(m)$ and the step-size $\lambda(m)$ increase, while the epoch length $N_e(m)$ decreases. A possible explanation for this behavior is that as $N_{\T}$ increases, the empirical Bellman operator $\widetilde{\T}$ becomes a more accurate estimation of the Bellman operator $\T$, which allows for a more aggressive step size and, consequently, a shorter epoch length. Moreover, it is worth noting that, in practice, the universal constants in the algorithm parameters may be conservative. Typically, smaller constants for epoch lengths and recentering sample sizes, or even a larger constant for the step size, can be employed (see Section~\ref{sec:simulation}).
\end{remark}

\begin{remark}[Geometric convergence with shorter epoch lengths]\label{Remark.2} Similar to the variance-reduced Q-learning~\cite{VRQ.Wainwright}, VRCQ exhibits a geometric convergence rate as a function of the epoch number $m$. VRCQ achieves this while utilizing only $N_e=\mathcal{O}(\frac{\log(\frac{D}{(1-\gamma)\delta})}{(1-\gamma)^2})$ iterations within the inner loop of each epoch. In contrast, variance-reduced Q-learning requires $N_e=\mathcal{O}(\frac{\log(\frac{D}{(1-\gamma)\delta})}{(1-\gamma)^3})$ iterations to achieve a similar outcome~\cite[Sec. 3.2]{VRQ.Wainwright}. Generally speaking, this feature is the primary contributing factor behind the improved sample complexity results presented in sections \ref{sec: Global.Minimax} and~\ref{sec: instance-dependent}.
\end{remark}

\subsection{Global Minimax Analysis of VRCQ}\label{sec: Global.Minimax}

In this section, we examine the worst-case sample complexity of VRCQ and establish a bound on the total number of samples required for VRCQ to return an $\epsilon$-accurate solution with high probability. For simplicity and consistency with the existing literature~\cite{Minimax.bound,VRQ.Wainwright,Q-learning-tight,MDVI}, we assume $\sigma_r=0$. First, it follows directly from Theorem \ref{theorem.1}.b that running VRCQ for $M=\log_\frac{1}{\phi} (\frac{\|\Theta^\star\|_{\infty}}{\epsilon})$ epochs results in a $\epsilon$-accurate solution with probability at least $1-\delta$. Moreover, the total number of samples is bounded by
\begin{align}
S=\sum_{m=0}^{M-1} \Big(N_e(m) + N_\T(m)\Big) \leq  c\Big(\frac{ M\log(\frac{M D}{(1-\gamma)\delta })}{(1-\gamma)^2} + \frac{\log(\frac{M D}{\delta})}{\phi^{2M}(1-\gamma)^2} \Big), \label{Eq:The sample complexity.b.1}
\end{align}
for some universal constant $c$. Taking to account that $\sup_{\Theta^\star \in \mathcal{M}(\Theta^\star,\gamma)} \|\Theta^\star\|_{\infty}\leq \frac{r_{\text{max}}}{1-\gamma}$, we have

\begin{align}\label{Eq: worst case bound Algorithm2.a}
S \leq c \Big(\frac{ \log(\frac{r_{\text{max}}}{(1-\gamma)\epsilon})\log\big(\frac{D}{(1-\gamma)\delta} \log(\frac{r_{\text{max}}}{(1-\gamma)\epsilon})  \big)}{(1-\gamma)^2} +\frac{r^2_{\text{max}} \log\big(\frac{D}{\delta} \log(\frac{r_{\text{max}}}{(1-\gamma)\epsilon})\big)}{{\epsilon^2(1-\gamma)^4}} \Big),
\end{align}
which do not mach the optimal cubic scaling in $\frac{1}{1-\gamma}$. Nevertheless, using a refined analysis, similar to the one done for the standard variance-reduced Q-learning \cite[Sec. 3.4.3]{SA.Wainwright}, we prove that VRCQ has the optimal sample complexity. At first, suppose that VRCQ is run for 
 $M_{\text{Init}}=\log_{\frac{1}{\phi}} (\frac{1}{\sqrt{1-\gamma}})$ epoches, then its output $\Theta_{M_{\text{Init}}}$ satisfies $\|\Theta_{M_{\text{Init}}}-\Theta^\star\|_{\infty} \leq \frac{r_{\text{max}}}{\sqrt{1-\gamma}}$ with high probability. Furthermore, based on \eqref{Eq: worst case bound Algorithm2.a}, the number of iterations is bounded by
\begin{align*}
S_{\text{Init}}  \leq c \Big(\frac{ \log(\frac{1}{(1-\gamma)})\log\big(\frac{D}{(1-\gamma)\delta} \log(\frac{1}{1-\gamma})  \big)}{(1-\gamma)^2} +\frac{\log\big(\frac{D}{\delta} \log(\frac{1}{(1-\gamma)})\big)}{{(1-\gamma)^3}} \Big).
\end{align*}

The following proposition, inspired by Proposition 1 in~\cite{VRQ.Wainwright}, states that running VRCQ from $\Theta_{M_{\text{Init}}}$ for a further 
logarithmic number of epochs results in an $\epsilon$-accurate solution.

\begin{proposition}[Minimax optimality of VRCQ]
\label{proposition.2} Suppose VRCQ is run from the initial point $\Theta_0= \Theta_{M_{\text{Init}}} $ for $M_{\text{Late}}=\log_\frac{1}{\phi} \Big(\frac{\Bar{c} r_{\text{max}} }{\sqrt{(1-\gamma)}\epsilon}\Big)$ iterations where $\Bar{c} \geq \frac{4\sqrt{2}\log(2)}{r_{\text{max}}}+1$, and the algorithm parameters are chosen according to Theorem \ref{theorem.1}.b. Then, the inequality $\|\Theta_M-\Theta^\star\|\leq \epsilon$ holds with probability at least $1-\delta$.
\end{proposition}

Note that, based on \eqref{Eq:The sample complexity.b.1}, the number of iterations required in Proposition \ref{proposition.2} is bounded by

\begin{equation*}
   S_{\text{Late}}  \leq c \Big(\frac{ \log(\frac{r_{\text{max}}}{\sqrt{1-\gamma}\epsilon})\log\big(\frac{D}{(1-\gamma)\delta} \log(\frac{r_{\text{max}}}{\sqrt{1-\gamma}\epsilon})  \big)}{(1-\gamma)^2} +\frac{r^2_{\text{max}}\log\big(\frac{D}{\delta} \log(\frac{r_{\text{max}}}{\sqrt{1-\gamma}\epsilon})\big)}{{\epsilon^2(1-\gamma)^3}} \Big).
\end{equation*}

As a result, the total number of iterations, counting both the initial iterations required to obtain $\Bar{\Theta}$ and later $S_{\text{late}}$ iterations, used to obtain this $\epsilon$-accurate solution is bounded by
\begin{equation}\label{Eq:VRCQ.minimax.bound}
 S= S_{\text{init}} + S_{\text{late}}  \leq  c \Big(\frac{ \log(\frac{r_{\text{max}}}{(1-\gamma)\epsilon})\log\big(\frac{D}{(1-\gamma)\delta} \log(\frac{r_{\text{max}}}{(1-\gamma)\epsilon})  \big)}{(1-\gamma)^2}+\frac{r^2_{\text{max}}\log\big(\frac{D}{\delta} \log(\frac{r_{\text{max}}}{(1-\gamma)\epsilon})\big)}{{\epsilon^2(1-\gamma)^3}} \Big).   
\end{equation}

The first term on the right-hand side of the above inequality represents the number of samples used as epoch lengths, i.e., $ \sum_{m} N_e (m)$, while the second term represents the number of samples used for recentring, i.e., $\sum_{m} N_\T (m)$.
As we can see, the second term is dominant, and it matches the lower bound $\Omega\Big(\frac{r^2_{\text{max}} \log(\frac{D} {\delta})}{\epsilon^2(1-\gamma)^3}\Big)$, up to a $\log\Big(\log(\frac{r_{\text{max}}}{(1-\gamma)\epsilon})\Big)$ factor. This additional logarithmic term results from using the union bound in the proof of Theorem \ref{theorem.1}.b.\par

\begin{remark}[VRCQ versus other algorithms: Minimax viewpoint]\label{Remark.3}
The worst-case sample complexity of various algorithms has been compiled in Table \ref{table:1}. As we can observe, Q-learning and Speedy Q-learning are suboptimal. Mirror Descent Value Iteration is minimax optimal up to some logarithmic factors; however, it requires $\epsilon$ to be sufficiently small when the discount factor $\gamma$ is close to $1$. The variance-reduced Q-learning has the following worst-case sample complexity~\cite[see][Sec. 3.4.3]{VRQ.Wainwright}
\begin{equation} \label{Eq:VR.minimax.bound}
S \leq c \Big(\frac{\log\left(\frac{r_{\text{max}}}{(1-\gamma)\epsilon}\right)\log\left(\frac{D}{(1-\gamma)\delta} \log\left(\frac{r_{\text{max}}}{(1-\gamma)\epsilon}\right)\right)}{(1-\gamma)^3} +\frac{r^2_{\text{max}}\log\left(\frac{D}{\delta} \log\left(\frac{r_{\text{max}}}{(1-\gamma)\epsilon}\right)\right)}{\epsilon^2(1-\gamma)^3}\Big).
\end{equation}

Similar to \eqref{Eq:VRCQ.minimax.bound}, the first term on the right-hand side of \eqref{Eq:VR.minimax.bound} represents the number of samples used as the epoch lengths, and the second term represents the number of samples used for recentring. A comparison between \eqref{Eq:VRCQ.minimax.bound} and \eqref{Eq:VR.minimax.bound} reveals that both VRCQ and variance-reduced Q-learning employ the same sample size for recentring. However, the number of samples used as epoch lengths in \eqref{Eq:VR.minimax.bound} is significantly larger than the corresponding term in \eqref{Eq:VRCQ.minimax.bound}. Consequently, when considered as functions of the horizon $(1-\gamma)^{-1}$, the right-hand side of \eqref{Eq:VR.minimax.bound} is dominated by its first term, resulting in the worst-case sample complexity of $\mathcal{O}\left(\frac{r_{\text{max}}^2\log\left(\frac{1}{1-\gamma}\right)\log\left(\frac{D}{(1-\gamma)\delta} \log\left(\frac{r_{\text{max}}}{(1-\gamma)\epsilon}\right)\right)}{\epsilon^2(1-\gamma)^3}\right)$. Thus, in the worst-case scenario, VRCQ outperforms variance-reduced Q-learning by a logarithmic factor in the discount complexity $(1-\gamma)^{-1}$. In the next section, we will demonstrate that by considering a stronger criterion for optimality, known as instance optimality, the distinction in performance between these two algorithms becomes more apparent.
\end{remark} 

\subsection{Instance-Dependent Analysis of VRCQ When $|\mathcal{U}|=1$}\label{sec: instance-dependent}
In this section, we explore the instance-dependent behavior of VRCQ in the non-asymptotic regime, where samples are limited. To simplify the analysis and facilitate the comparison between our algorithm and others~\cite{Khamaru.TD,Khamaru.QL, Root.SA, Q-learning-PL, Pananjady}, we focus on the specific scenario where the action space contains only a single action, i.e., $|\mathcal{U}|=1$. In this case, the problem of estimating the optimal Q-function coincides with the policy evaluation problem. Moreover, a given problem instance can be characterized by the pair $\mathcal{P}=(\mathbb{P}, r)$ along with a discount factor $\gamma$, where $r\in \mathbb{R}^D$ represents the reward vector and $\mathbb{P} \in [0, 1]^{D \times D}$ denote a row-stochastic (Markov) transition matrix. The value function of the problem instance $\mathcal{P}$ (denoted here by the vector $\Theta^\star (\mathcal{P}) \in \mathbb{R}^D$) is the sum of the infinite-horizon discounted rewards. This value function is the unique fixed point of the Bellman operator $T(\Theta)=r + \gamma \mathbb{P} \Theta $. As discussed in Section \ref{sec:background}, in the generative setting, at each iteration, we have access to samples $(\widehat{\mathbb{P}}_n, \hat{r}_n)$, where $\hat{r}_n$ is the noisy observation of the reward, and $\widehat{\mathbb{P}}_n$ denotes a draw of a random matrix with $\{0, 1\}$ entries and a single one in each row. The empirical Belman operator is $\widehat{\T}_n(\Theta)=\hat{r}_n + \gamma \widehat{\mathbb{P}}_n \Theta $.

\textbf{The local minimax risk and complexity measures.} Suppose we have $N$ i.i.d samples $\{(\widehat{\mathbb{P}}_n, \hat{r}_n)\}_{i=1}^{N}$ from our observation model. The \emph{local non-asymptotic minimax risk} for $\Theta^\star(\mathcal{P})$ at an instance $\mathcal{P}=(\mathbb{P},r)$ is defined as~\cite{Khamaru.TD}
\begin{equation}
    \mathcal{M}_{N}(\mathcal{P})=\sup_{\Bar{\mathcal{P}}}  \inf_{\widehat{\Theta}_N} \max_{\mathcal{Q}\in \{\mathcal{P}, \Bar{\mathcal{P}}\}} \sqrt{N} \mathbb{E}_Q \|\widehat{\Theta}_N-\Theta^\star(\mathcal{Q}) \|_{\infty},
\end{equation}
 where the infimum is taken over all estimators $\widehat{\Theta}_N$ that are measurable functions of $N$ i.i.d observations. Moreover, For a given instance $\mathcal{P}=(\mathbb{P},r)$ we define the complexity measures
 \begin{align*}
    v(\mathcal{P} ) &:= \Big\| \text{diag}\Big( (I-\gamma \mathbb{P})^{-1} \text{cov}_{\widehat{\mathbb{P}} \sim \mathbb{P}  }\big( \widehat{\mathbb{P}} \Theta^\star (\mathcal{P}) \big) (I-\gamma \mathbb{P})^{-T}  \Big) \Big\|_{\infty}^{\frac{1}{2}},\\
    \rho(\mathcal{P} ) &:=\sigma_r \Big\| \text{diag}\Big( (I-\gamma \mathbb{P})^{-1} (I-\gamma \mathbb{P})^{-T}  \Big) \big\|_{\infty}^{\frac{1}{2}}.
\end{align*}
The following theorem lower bounds $\mathcal{M}_{N}(\mathcal{P})$ using the complexity measures $v (\mathcal{P})$ and $\rho(\mathcal{P})$.

\begin{theorem}[~\cite{Khamaru.TD} Lower bound on $\mathcal{M}_{N}(\mathcal{P})$]
 \label{theorem.2}
 There exists a universal constant $c\geq 0$ such that for any instance $\mathcal{P}$, the local non-asymptotic minimax risk is lower bounded as 
\begin{equation*}
      \mathcal{M}_{N}(\mathcal{P}) \geq c\big( \gamma v (\mathcal{P})+\rho(\mathcal{P}) \big).
\end{equation*}
This bound is valid for all sample sizes $N$ that satisfy
\begin{equation} \label{Eq: Theorem3. sample size}
    N\geq N_0 : = \frac{ \max \{ \gamma^2, \frac{\|\Theta^\star(\mathcal{P})\|^2_{\text{span}}}{v^2(\mathcal{P})} \}}{(1-\gamma)^2}.
\end{equation}\par
\end{theorem}

The statement of Theorem \ref{theorem.2} indicates that the local complexity of estimating the value function $\Theta^\star(\mathcal{P})$ induced by the instance $\mathcal{P}$ is governed by the quantities $v(\mathcal{P})$ and $\rho(\mathcal{P})$. The minimum sample size condition is natural since when the rewards are observed with noise (i.e., for any $\sigma_r > 0$), this condition is necessary to obtain an estimate of the value function with $\mathcal{O}(1)$ error (see~\cite{Khamaru.TD,Pananjady} for more details).
 
The next theorem provides an upper bound on the performance of VRCQ in terms of the local complexity measures $v(\mathcal{P})$ and $\rho (\mathcal{P})$. Similar to Theorem \ref{theorem.2}, we provide this bound on the expected error. A high-probability bound can also be derived by selecting the algorithm parameters according to Theorem \ref{theorem.1}.b and following a similar line of reasoning.

\begin{theorem}[Non-asymptotic optimality of VRCQ] \label{theorem.3}
Suppose that the input parameters of Algorithm \ref{algorithm.2} are chosen according to Theorem \ref{theorem.1}.a. Moreover, suppose that the total sample size $N$ satisfies 
\begin{equation}\label{Eq: Theorem4.sample size}
N\geq\frac{ c \gamma \log( D)}{(1-\gamma)^2},
\end{equation}
where $c$ is a sufficiently large universal constant. Then, by running Algorithm \ref{algorithm.2} from any initial point $\Theta_0$ for $M=\log_{\frac{1}{\phi}}({\frac{\sqrt{1-\phi^2} (1-\gamma)\sqrt{N}} {8\sqrt{\gamma \log 2D}} })$
epochs, the resulting output $\Theta_M$ satisfies
\begin{align}
   \mathbb{E} \|\Theta_M-\Theta^\star ( \mathcal{P} ) \|_{\infty} \leq  &\Big(\frac {8 \sqrt{ \gamma \log (2D) }} {\sqrt{N}\sqrt{1-\phi^2} (1-\gamma)}\Big)^{1+\log_{\frac{1}{\phi}} (\frac{9}{6+\phi}) }  \|\Theta_0-\Theta^\star (\mathcal{P} )\|_{\infty}+\frac{2\|\Theta^\star (\mathcal{P})\|_{\text{span}} \log(2 D) } {(\frac{4}{3}-\frac{1}{\phi})(1-\phi^2) (1-\gamma) N }  \nonumber\\ \label{Eq: Theorem4.Upper bound}
&+ \frac{13}{\sqrt{1-\phi^2} }   \big( \rho(\mathcal{P})+\gamma v(\mathcal{P}) \big) \sqrt{\frac{\log 2D}{N}}.    
\end{align}\par
\end{theorem}

\begin{remark}[Instance-dependent upper and lower bounds]\label{remark.4}
The first term on the right-hand side of the upper bound \eqref{Eq: Theorem4.Upper bound} depends on the initialization $\Theta_0$. When viewed as a function of the sample size $N$, this initialization-dependent term can be made smaller than other terms by choosing a suitable convergence rate $\phi$. For instance, by setting $\phi=0.875$, we have ($\log_{\frac{1}{\phi}}(\frac{9}{6+\phi})\geq 2$). It should be noted that a careful look at the proof of Theorem \ref{theorem.3} reveals that another way to make the initialization-dependent term small is by increasing the epoch length $N_e$ by a constant factor. This indicates that the performance of VRCQ does not depend on the initialization $\Theta_0$ in a significant way. The second and the third terms are the dominating terms. Furthermore, assuming that the minimum sample size requirement \eqref{Eq: Theorem4.sample size} is met, the upper bound matches the lower bound up to a logarithmic term in the dimension. Similarly, up to a logarithmic factor in dimension, the minimum sample size requirement in Theorem \ref{theorem.3} matches the lower bound \eqref{Eq: Theorem3. sample size}.\par
   
\end{remark}

\begin{remark}[VRCQ versus other algorithms: Instance-dependent behavior]\label{remark.5} The instance-dependent guarantees for various algorithms have been collected in Table \ref{table:2}. In~\cite{Khamaru.TD}, it is shown that while Q-learning with PR averaging is instance-optimal in the asymptotic setting (i.e., when the sample size $N$ converges to infinity), it fails to achieve the correct rate in the non-asymptotic regime, even when the sample size is quite large (see also~\cite[Sec. 5]{Q-learning-PL} for further discussions about the instance optimally of Q-learning with PR averaging). Moreover,~\cite{Khamaru.TD} demonstrate that the variance-reduced Q-learning proposed in~\cite{VRQ.Wainwright} is instance optimal when sample size satisfies the condition $N  \geq \frac{c\log(D) }{(1-\gamma)^3}$. This sample size requirement, unlike VRCQ, does not match the lower bound \eqref{Eq: Theorem3. sample size}. Another notable algorithm is Root-SA proposed in~\cite{Root.SA}. This algorithm provides an instance-dependent bound on the span seminorm of the error, assuming that the sample size meets the condition $N \geq \frac{c\log(\frac{D}{1-\gamma})}{(1-\gamma)^2}$~\cite[Corollary 6]{Root.SA}. Nevertheless, since \(\frac{1}{2}\|\cdot \|_{\text{span}} \leq \| \cdot \|_\infty\), meaning the span seminorm is dominated by the \(\ell_\infty\)-norm, bounds on the span seminorm of the error are inherently weaker than those on the \(\ell_\infty\)-error. Therefore, since the result is not expressed in the $\ell_\infty$-norm, we have not included Root-SA in Table \ref{table:2}.
\end{remark}

\section{Numerical Results} \label{sec:simulation}\par
In this section, we validate our results through two numerical simulations: one demonstrating the instance optimality of VRCQ, and the other illustrating its performance on random Garnet MDPs~\cite{Garnet.1,Garnet.2}.

\begin{figure}
    \centering
    \includegraphics[width=5.5cm]{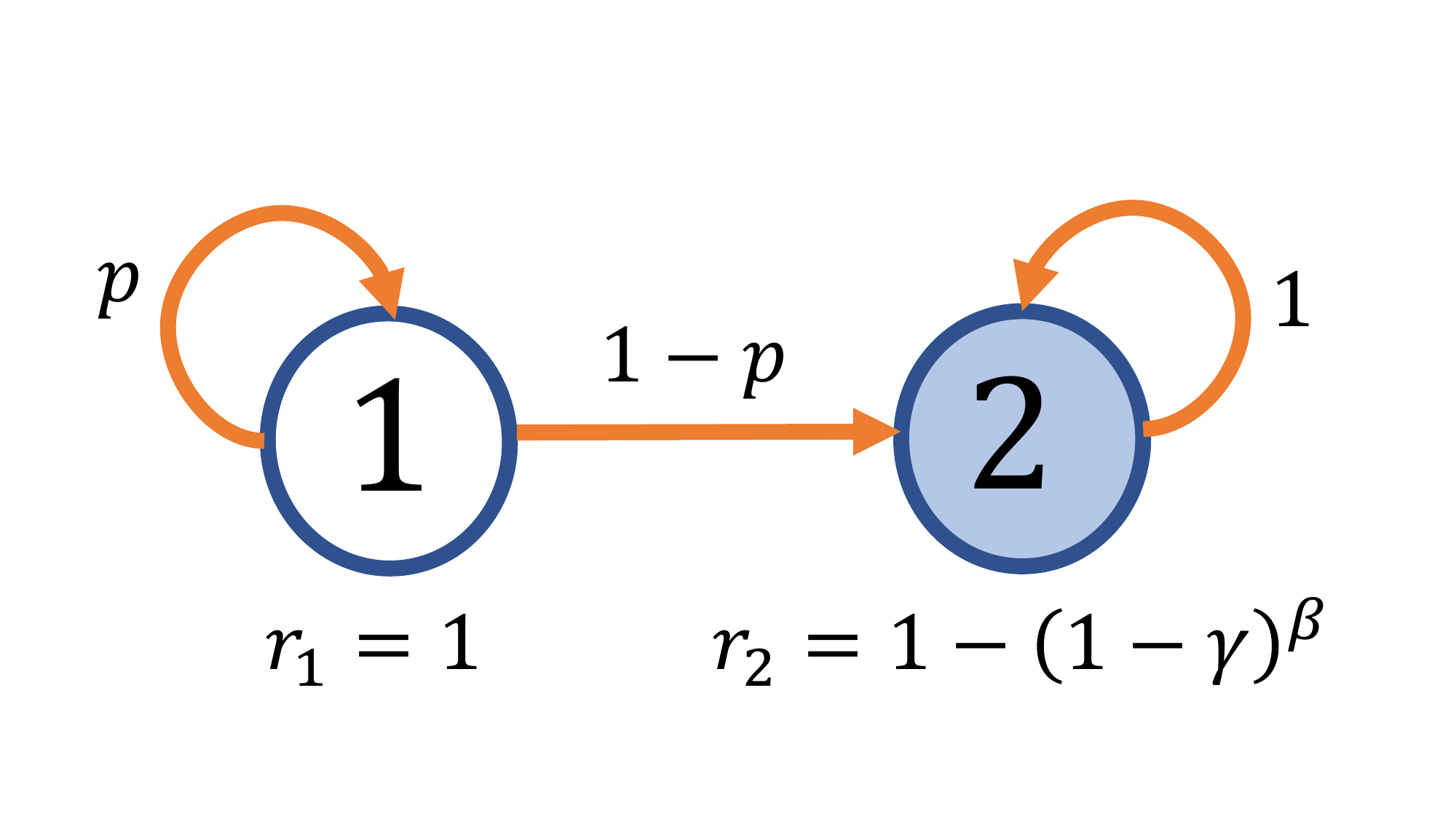}
    \caption{Transition diagram a class of MDP, adopted from~\cite{Khamaru.TD}. The scalers $\beta \geq 0$, and $ 0 < p < 1 $ are parameters of the construction. The chain remains in state $1$ with probability $p$ and transitions to state $2$ with probability $1-p$. State 2 is absorbing. }
    \label{fig:MDP}
\end{figure}

\textbf{Example 1} (Instance optimality)\textbf{.} Consider the $2$-state MDP illustrated in Figure \ref{fig:MDP}, where \( p = \frac{4\gamma - 1}{3\gamma} \). This sub-family of MDPs is fully parameterized by the pair \((\gamma, \beta)\). Assuming \(\sigma_r = 0\), straightforward calculations show that \( v(\mathcal{P}) = \frac{c}{(1 - \gamma)^{1.5 - \beta}} \) and \( \rho(\mathcal{P}) = 0 \), where \( c \) is a constant. Hence, according to Theorem \ref{theorem.2} the local minimax risk satisfies
\begin{equation}\label{Eq:Example.lower bound}
\mathcal{M}_{N}(\mathcal{P}) \geq \frac{c}{(1-\gamma)^{1.5-\beta}}.    
\end{equation}
For numerical results, we generate a range of MDPs with different discount factors $\gamma \in [0.96, 0.997]$, keeping the value of $\beta$ fixed. For each $\gamma$, we consider the problem of estimating $\Theta^\star(\mathcal{P})$ using
\begin{equation}\label{Eq:Example.N}
N=\frac{100}{(1-\gamma)^2}    
\end{equation}
samples. It follows directly from \eqref{Eq:Example.lower bound} and \eqref{Eq:Example.N} that for an instance-optimal algorithm, a linear relationship is expected between the log $\ell_\infty$-error and the log complexity $\frac{1}{1-\gamma}$, with a slope of $\frac{1}{2}-\beta$. 

\begin{figure}
     \begin{subfigure}[b]{0.345\textwidth}
         \centering
    \includegraphics[width=\textwidth]{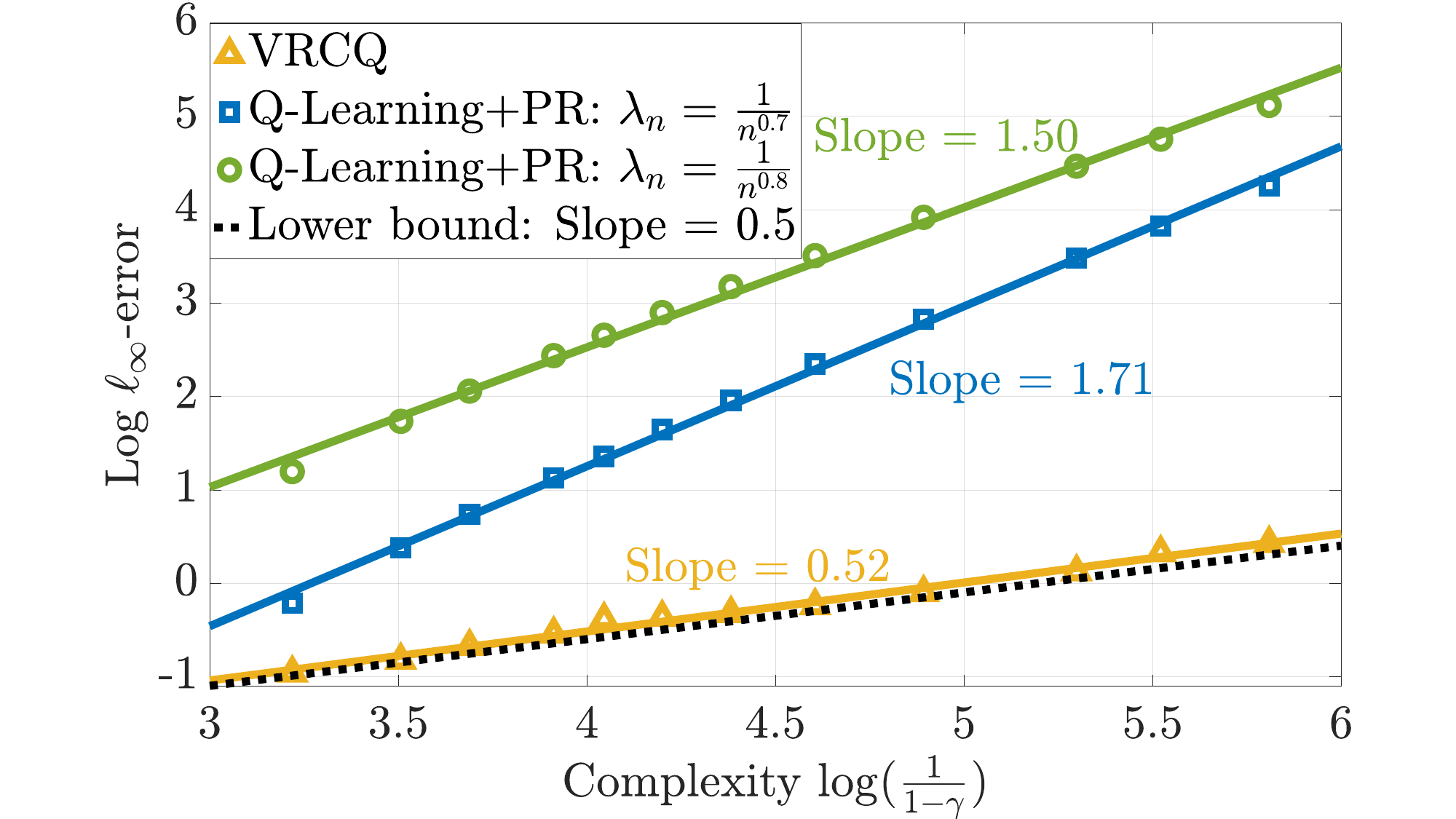}
         \caption{ $\beta=0$}
     \end{subfigure}
     \hspace{-0.55cm}
      \begin{subfigure}[b]{0.345\textwidth}
         \centering
    \includegraphics[width=\textwidth]{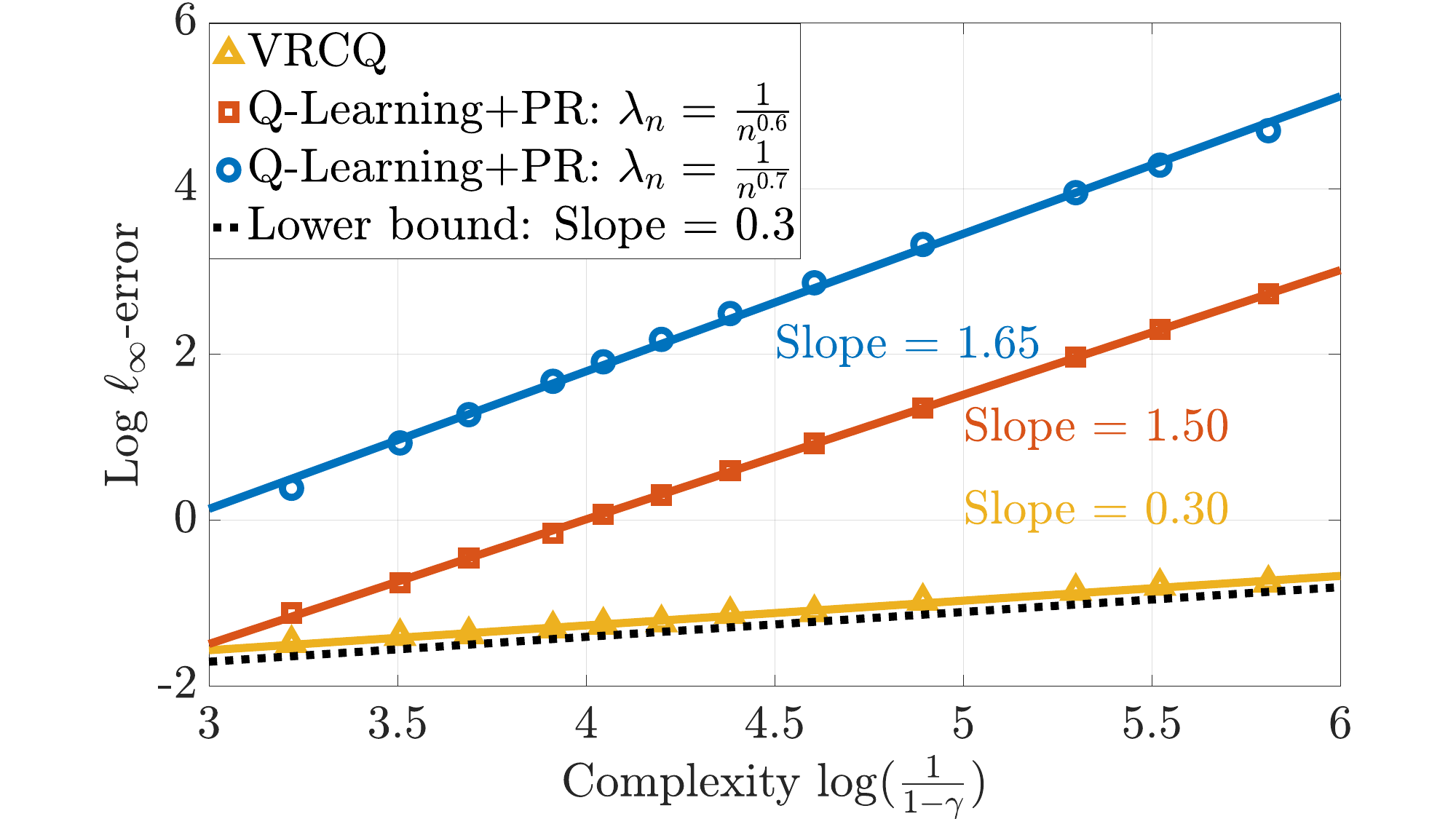}
         \caption{$\beta=0.2$}
     \end{subfigure} 
        \hspace{-0.55cm}
  \begin{subfigure}[b]{0.345\textwidth}
         \centering
         \includegraphics[width=\textwidth]{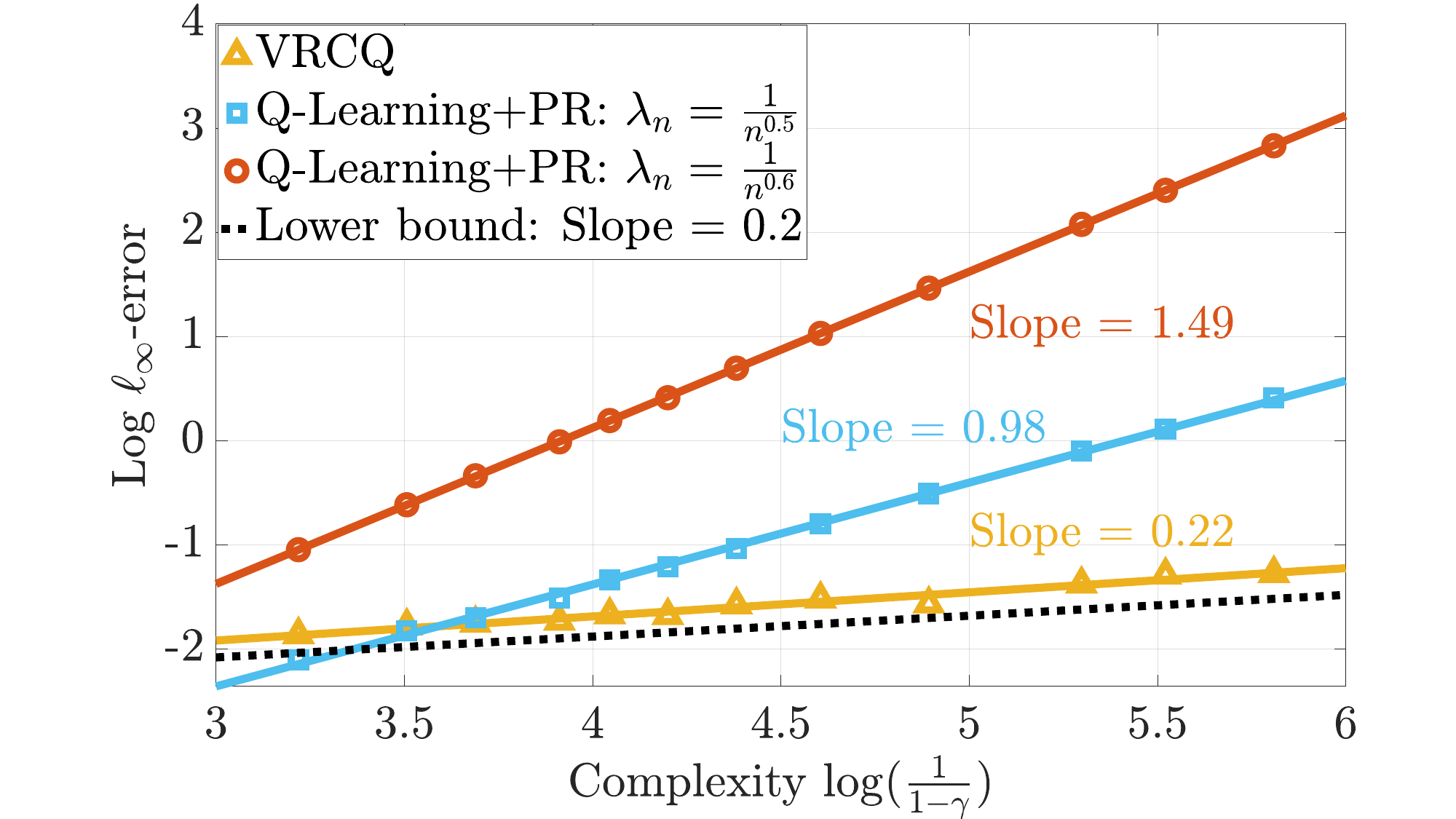}
         \caption{$\beta=0.3$}
      
     \end{subfigure} 
     
     \caption{ Log-log plots of the $\ell_\infty$-error versus complexity parameter $\frac{1}{1-\gamma}$ for
                different algorithms. Each data point is an average of $500$ independent trials.}
            \label{fig:Slope}
\end{figure}

Figure \ref{fig:Slope} shows log-log plots of the $\ell_\infty$-error as a function of the complexity parameter $\frac{1}{1-\gamma}$ for VRCQ (yellow) and PR averaged Q-learning with four different step sizes, $\lambda_n = n^\eta$, where $\eta \in \{-0.8, -0.7, -0.6, -0.5\}$, along with the theoretical local lower bound (dashed). Each data point is obtained by averaging $500$ independent trials. For VRCQ, we utilize the following parameters: $M=15$, $\phi=0.95$, $N_\T(m)=\frac{0.738}{\phi^{2m}(1-\gamma)^2}$, and $N_{e}(m)=\frac{5}{(1-\gamma)^2}$. Note that we have $\sum_{m=0}^{14} N_\T(m)+N_e(m) \leq N$. Moreover, constant factors in $N_e$ and $N_\T$ are smaller than those suggested by Theorem \ref{theorem.1}.a. Nevertheless, as shown in Figure~\ref{fig:Slope}, we observe that VRCQ achieves the instance-optimal rate. In contrast, PR averaged Q-learning with polynomial step size fails to achieve the correct rate, even though it is optimal for sufficiently large sample sizes~\cite[Theorem 5.1]{Q-learning-PL}. Finally, it is worth noting that the variance-reduced Q-learning is not applicable in this setting because it requires $N_e=\mathcal{O}\left(\frac{1}{(1-\gamma)^3}\right)$ samples as the epoch length whereas in this problem, only $\mathcal{O}\left(\frac{1}{(1-\gamma)^2}\right)$ samples are available~\cite[see][Theorem 3.3] {Khamaru.TD}.\par

\par

\textbf{Example 2} (Performance on random Garnet MDPs)\textbf{.} To compare the practical performance of VRCQ with the standard variance-reduced Q-learning~\cite{VRQ.Wainwright,Khamaru.QL,Khamaru.TD}, we consider the problem of estimating the optimal Q-function of randomly generated Garnet MDPs~\cite{Garnet.1,Garnet.2}. A Garnet MDP is characterized by three integer parameters: (\romannumeral 1 ) the number of states $|\mathcal{X}|$, (\romannumeral 2) the number of actions $|\mathcal{U}|$, and (\romannumeral 3) the branching factor $b$, specifying the number of possible next states for each state-action pair. The next states are selected randomly from the state set without replacement. Moreover, the probability of going to each next state is generated by partitioning the unit interval at $b-1$ randomly chosen cut points between $0$ and $1$. 

\begin{figure} 
     \begin{subfigure}[b]{0.33\textwidth} 
         \centering         \includegraphics[width=\textwidth]{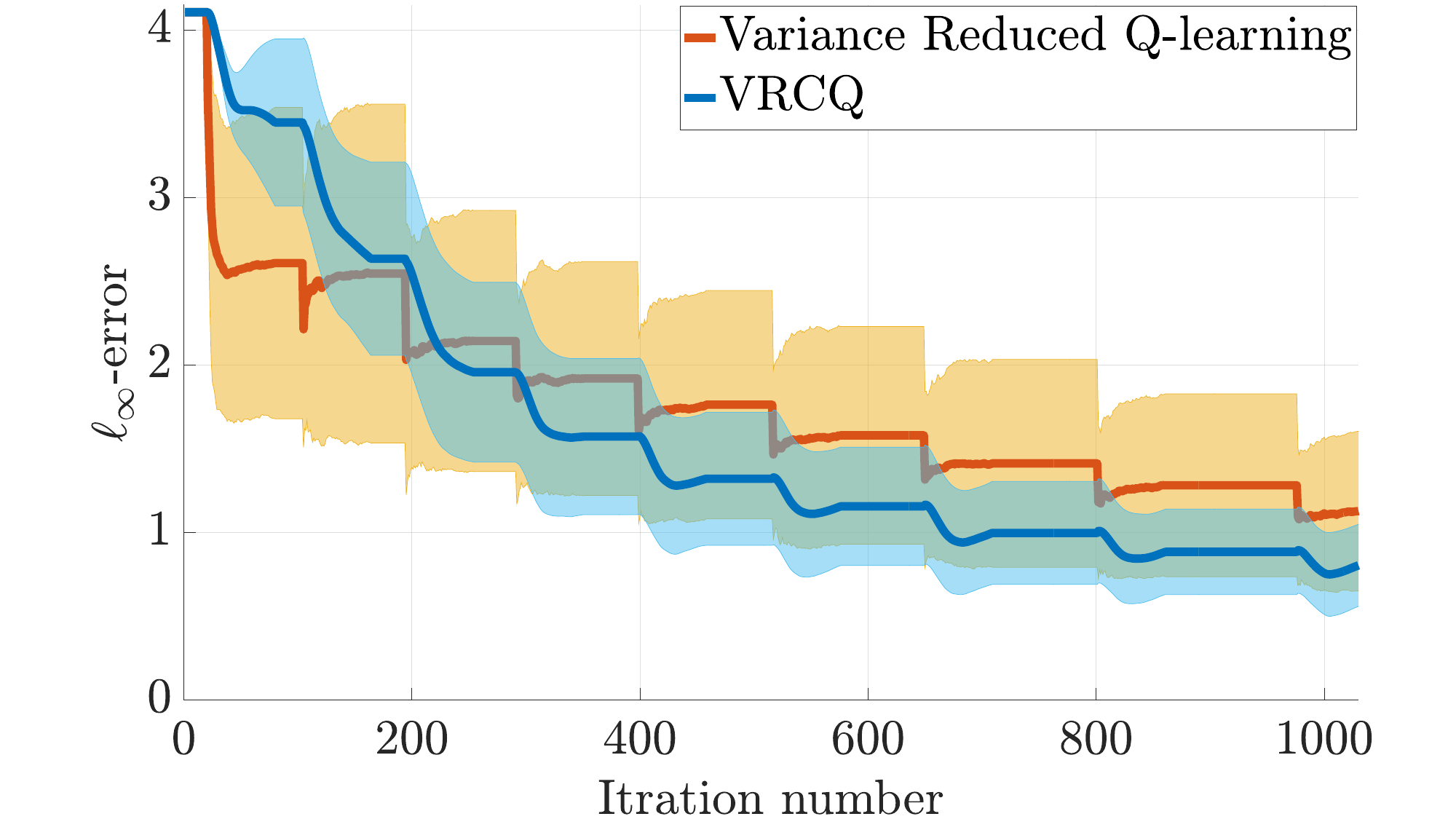}
         \caption{$\gamma=0.80$}
     \end{subfigure}
     \hspace{-0.3cm}
      \begin{subfigure}[b]{0.33\textwidth}
         \centering
         \includegraphics[width=\textwidth]{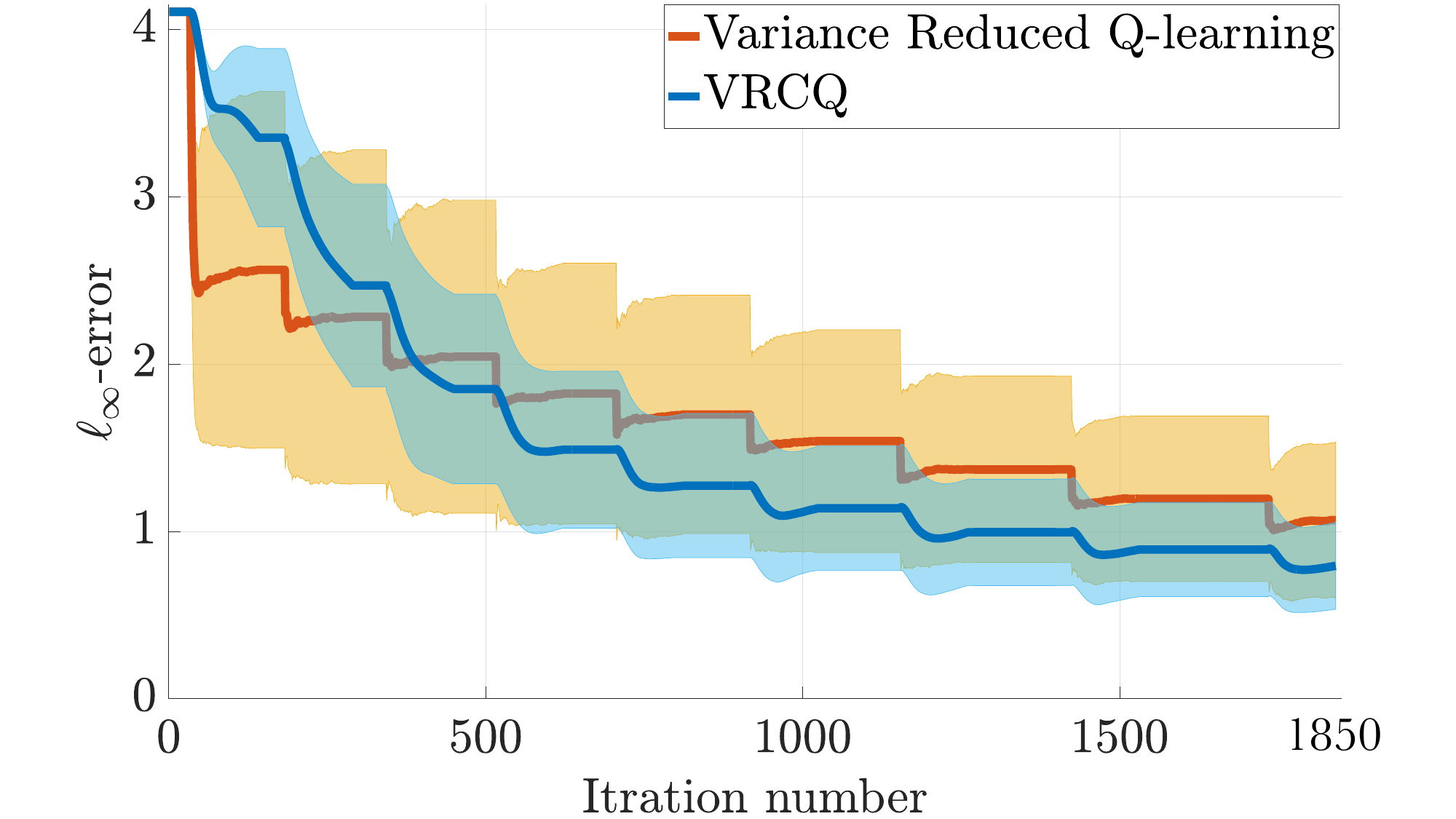}
         \caption{$\gamma=0.85$}
     \end{subfigure} 
     \hspace{-0.3cm}
       \begin{subfigure}[b]{0.33\textwidth}
         \centering
         \includegraphics[width=\textwidth]{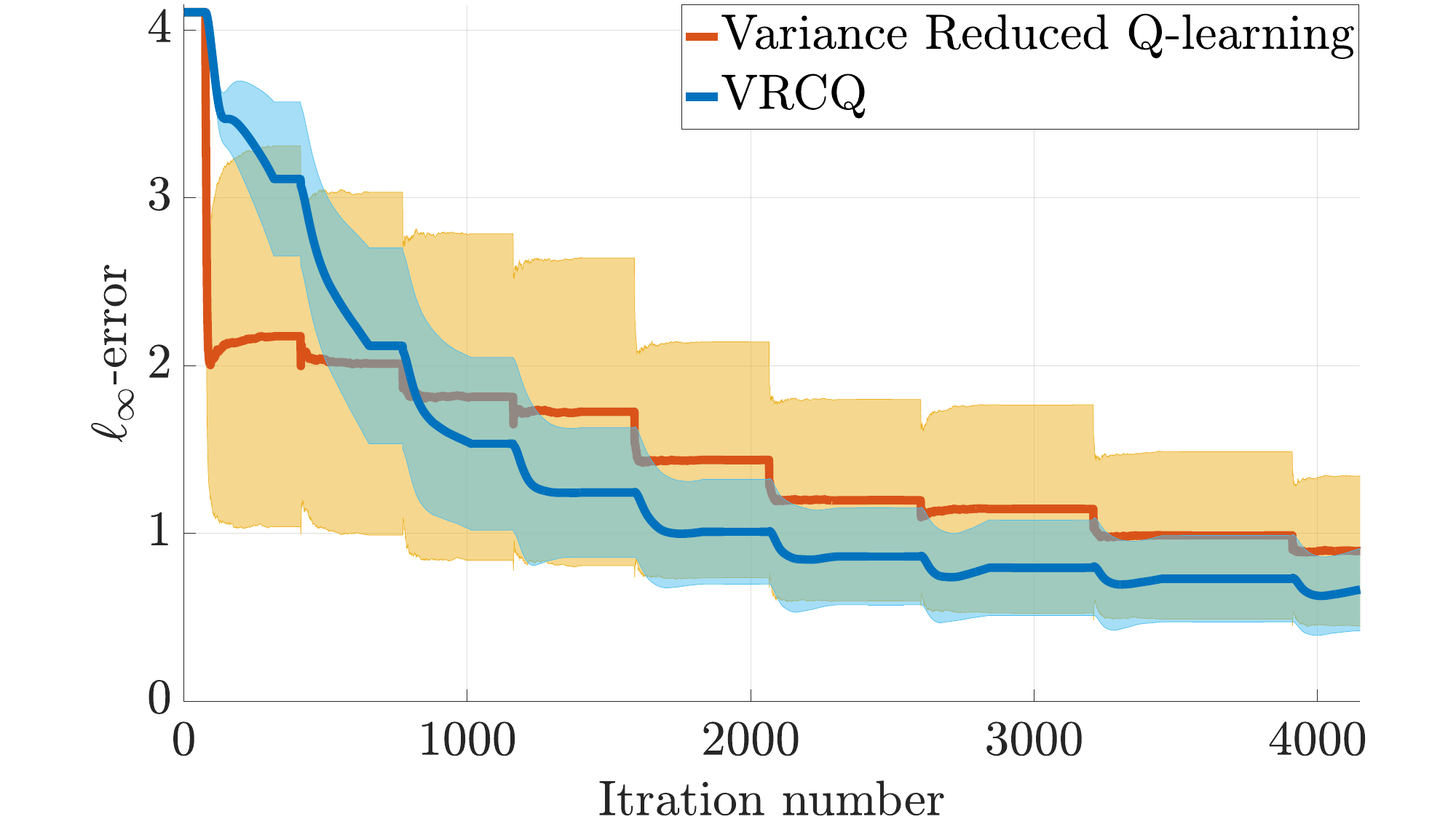}
         \caption{$\gamma=0.90$}
     \end{subfigure} 
     \caption{ Comparison of the convergence behavior of VRCQ and variance-reduced Q-learning. For a given algorithm and value of $\gamma$, we run the algorithm for a certain number of epochs, thereby obtaining a path of $\ell_\infty$-errors at each iteration. We averaged these paths over a total of $500$ independent trials. The radius of the shaded area at each iteration represents the standard deviation of the $\ell_\infty$-error.}
     \label{fig.garnet}
\end{figure}

In our experiments, we set $|\mathcal{X}|=20$, $|\mathcal{U}|=2$, and $b=2$. We implement VRCQ and variance-reduced Q-learning using identical recentering sample sizes and epoch lengths. Figure \ref{fig.garnet} plots the $\ell_\infty$-error of both algorithms versus the number of iterations for three different discount factors. As we can see, in the early iterations, when the error due to initialization is typically larger than the error due to noise fluctuations in the algorithm, the averaged error of variance-reduced Q-learning decreases faster compared to VRCQ. Nevertheless, after a certain number of iterations, VRCQ outperforms variance-reduced Q-learning by achieving a lower averaged error and variance.

\section{Conclusion and Future Directions}\label{sec: Conclusion}\par
We studied the problem of estimating the optimal Q-function for $\gamma$-discounted MDPs in the synchronous setting. We introduced a novel model-free algorithm, VRCQ, and demonstrated that it is not only minimax optimal, but also, when the action set is a singleton, it achieves non-asymptotic instance optimality while requiring the theoretically minimum number of samples.  We conclude this article with the following potential future directions. \par
(\romannumeral 1) \textbf{Generalization to cone-contractive operators.} Although our findings were primarily presented within the Q-learning framework, many techniques and results in this study extend beyond the Q-learning problem. For example, by adapting a methodology similar to \cite{SA.Wainwright}, Algorithm \ref{algorithm.1}, and Proposition \ref{proposition.1} can be generalized to the broader task of finding the fixed point of an operator that is monotone and quasi-contractive with respect to an underlying cone.\par

(\romannumeral 2) \textbf{Studying the instance-dependent behavior of VRCQ for $|\mathcal{U}| > 1 $.} Recently,~\cite{Khamaru.QL} provided an instance-dependent lower bound similar to Theorem 3 for MDPs with $|\mathcal{U}| > 1$, assuming that the sample size scales as $(1-\gamma)^{-2}$. Given that VRCQ is already instance-optimal for this sample size when $|\mathcal{U}| = 1$, we conjecture that it remains optimal for $|\mathcal{U}| > 1$. Proving this conjecture is an interesting direction for future research, especially considering that, to the best of our knowledge, none of the existing algorithms in the literature matches this lower bound, e.g., Root-SA is instance optimal when the sample size scales as $(1-\gamma)^{-4}$~\cite[Corollary 5]{Root.SA}, while variance-reduced Q-learning, under certain conditions, is instance optimal when the sample size scales as $(1-\gamma)^{-3}$~\cite [Theorem~2] {Khamaru.QL}.

 (\romannumeral 3) \textbf{Generalization to other RL settings.} Another potential future direction is to explore the applicability of our proposed methods in other reinforcement learning settings, such as the episodic setting~\cite{Bubek, azar.regret}, the asynchronous setting~\cite{Li.Asynchronous, Asynchronous.2}, and Q-learning with function approximation in large (possibly continuous) state-action space~\cite{Meyn, wang}.

\section{Proofs of the Main Results} \label{sec: proofs}
\subsection{Preliminaries}\label{sec: proofs.pre}

In this section, we provide the proofs of the theorems and propositions presented in the paper. These proofs are based on the following preparatory lemmas. The first lemma (Lemma \ref{lemma.1}) establishes a non-asymptotic upper bound on the performance of cascade Q-learning for a given sequence of $\gamma$-contractive operators.
\par
\begin{lemma}\label{lemma.1}
     Consider a sequence of operators $\{\widehat{H}_n: \mathbb{R}^{|\mathbb{X}|\times|\mathbb{U}|} \rightarrow \mathbb{R}^{|\mathbb{X}|\times|\mathbb{U}|} \}_{n=1}^{N_e}$ that are $\gamma$-contractive in the $\ell_\infty$-norm and the sequences $\{Y_{n}\}_{n=1}^{N_e}$ and $\{Z_{n}\}_{n=1}^{N_e}$ generated by the recursions
\begin{subequations}
     \begin{align}
     Y_{n+1}&=(1-\lambda) Y_{n} + \lambda Z_{n} \label{Eq:Lemma1.Y}, \\ 
     Z_{n+1}&=(1-\lambda) Z_{n} + \lambda \widehat{H}_n (Y_{n+1}) \label{Eq:Lemma1.Z},
    \end{align}
\end{subequations}
with the step size $\lambda\in (0,1)$ and initial conditions $Z_1=Y_1=\Theta_0$. Then, for any $\Theta_H \in \mathbb{R}^{|\mathbb{X}|\times|\mathbb{U}|} $, we have
\begin{equation}\label{Eq: lemma 1. main}
     \Big\|\frac{1}{N_e}\sum_{n=1}^{N_e}Y_{n+1} -\Theta_H\Big\|_{\infty}\leq  \frac{2\|\Theta_0-\Theta_H\|_{\infty}}{(1-\gamma)\lambda N_e} + \frac{1}{1-\gamma}\left( \frac{1-\lambda}{N_e} \sum_{n=1}^{N_e} \| P_n^y \|_{\infty}+ \frac{\lambda}{N_e} \sum_{n=1}^{N_e} \| P_n^z \|_{\infty}\right),
\end{equation}
where the random matrix sequences $\{P_n^y\}_{n=1}^{N_e}$ and $\{P_n^z\}_{n=1}^{N_e}$ are defined via the recursions
\begin{subequations}
\begin{align}\label{Eq:recursion.Py}
    P^y_{n+1}&= (1-\lambda) P^y_n+\lambda P_n^z, \\
    P^z_{n+1}&=(1-\lambda)P^z_n+\lambda W_n, \label{Eq:recursion.Pz}\\
    W_n&=\widehat{H}_n(\Theta_H)-\Theta_H,
\end{align}
\end{subequations}
and initialized at $P_1^z=P_1^y=0$.
\end{lemma}
\par

The next lemma provides upper bounds, both in expectation and with high probability, over the expression $\frac{1-\lambda}{N_e}\sum_{n=1}^{N_e} \|P_n^y\|_{\infty} + \frac{\lambda}{N_e}\sum_{n=1}^{N_e} \|P_n^z\|_\infty$, the second term on the right-hand side of \eqref{Eq: lemma 1. main}, in some scenarios.

\begin{lemma}\label{lemma.2}
  Consider the stochastic processes \eqref{Eq:recursion.Py} and \eqref{Eq:recursion.Pz} with stepsize $\lambda=\frac{1}{\sqrt{N_e}}$. We denote the fixed points of the Bellman operator $\T$ and the operator $H(\Theta):=\T(\Theta)-\T(\Theta_m)+ \widetilde{\T} (\Theta_m)$ by $\Theta^\star$ and $\widehat{\Theta}$, respectively.
\begin{enumerate}[label=(\alph*)]
    \item Suppose $\widehat{H}_n=\widehat{\T}_n$, and $\Theta_H=\Theta^\star$, then we have
    \small
    \begin{subequations}
    \begin{align}
      &\frac{1-\lambda}{N_e}\sum_{n=1}^{N_e}\mathbb{E}\|P_n^y \|_{\infty}+ \frac{\lambda}{N_e}\sum_{n=1}^{N_e}\mathbb{E}\|P_n^z \|_{\infty} \leq \frac{2\gamma}{3}\lambda^2 \log(2D) \|\Theta^\star\|_{\text{span}}+2\lambda\sqrt{2\log(2D)} \big(\| \sigma (\Theta^\star)\|_{\infty}+\sigma_r\big), \label{Eq:Lemma2.a.1}   \\ 
     &\frac{1-\lambda}{N_e}\sum_{n=1}^{N_e} \|P_n^y \|_{\infty}+ \frac{\lambda}{N_e}\sum_{n=1}^{N_e} \|P_n^z \|_{\infty} \leq \frac{2\gamma}{3}\lambda^2 \log(\frac{8 N_e D}{\delta}) \|\Theta^\star\|_{\text{span}}+2\lambda\sqrt{2\log(\frac{8 N_e D}{\delta})} \big(\| \sigma (\Theta^\star)\|_{\infty}+\sigma_r\big), \label{Eq:Lemma2.a.2}
    \end{align}
    \end{subequations}
    \normalsize
    with probability at least $1-\delta$.
    \item Suppose $\widehat{H}_n (\Theta)= \widehat{\T}_n (\Theta)-\widehat{\T}_n(\Theta_m)+ \widetilde{\T} (\Theta_m)$ and $\Theta_h=\Theta^\star$. Define the function $C(x,\delta)=\frac{2x}{3}\log(\frac{2D}{\delta})+\sqrt{2x\log(\frac{2D}{\delta})}$, we have
    \begin{subequations}
 \begin{align}
     \frac{1-\lambda}{N_e}\sum_{n=1}^{N_e} \mathbb{E}\|P_n^y \|_{\infty}+ \frac{\lambda}{N_e}\sum_{n=1}^{N_e} \mathbb{E}\|P_n^z \|_{\infty}  \leq& \big(2\gamma C(\lambda^2, 1 ) + \gamma C(\frac{1}{N_\T}, 1)\big) \|\Theta_m-\Theta^\star\|_{\infty}  \nonumber \\
     &+\gamma C(\frac{1}{N_\T},1)\|\Theta^\star\|_{\infty}+\sqrt{\frac{2\log(2D)}{N_\T}} \sigma_r, \label{Eq:Lemma2.b.1}  \\
    \frac{1-\lambda}{N_e}\sum_{n=1}^{N_e} \|P_n^y \|_{\infty}+ \frac{\lambda}{N_e}\sum_{n=1}^{N_e} \|P_n^z \|_{\infty}  \leq& \big(2\gamma C(\lambda^2, \frac{\delta}{5 N_e M} ) + \gamma C(\frac{1}{N_\T}, \frac{\delta}{5 M})\big) \|\Theta_m-\Theta^\star\|_{\infty} \nonumber \\
    &+\gamma C(\frac{1}{N_\T},\frac{\delta}{5M})\|\Theta^\star\|_{\infty}+ \sqrt{\frac{2\log(\frac{10 M D}{\delta})}{N_\T}} \sigma_r,  \label{Eq:Lemma2.b.2} 
    \end{align}
    \end{subequations}
with probability at least $1-\frac{\delta}{M}$.
\item Suppose $\widehat{H}_n (\Theta)= \widehat{\T}_n (\Theta)-\widehat{\T}_n(\Theta_m)+ \widetilde{\T} (\Theta_m)$ and $\Theta_h=\widehat{\Theta}$, then we have
\begin{subequations}
\begin{align}\label{Eq: Lemma2.c.1}
    &\frac{1-\lambda}{N_e}\sum_{n=1}^{N_e} \mathbb{E}\|P_n^y \|_{\infty}+ \frac{\lambda}{N_e}\sum_{n=1}^{N_e} \mathbb{E}\|P_n^z \|_{\infty} \leq  2\gamma C(\lambda^2,1) \|\Theta_m-\widehat{\Theta}\|_{\infty},\\     
     &\frac{1-\lambda}{N_e}\sum_{n=1}^{N_e} \|P_n^y \|_{\infty}+ \frac{\lambda}{N_e}\sum_{n=1}^{N_e} \|P_n^z \|_{\infty} \leq  2\gamma C(\lambda^2,\frac{\delta}{5 M N_e}) \|\Theta_m-\widehat{\Theta}\|_{\infty}\label{Eq: Lemma2.c.2},
\end{align}
\end{subequations}
with probability at least $1-\frac{2\delta}{5M}$.
\end{enumerate}\par

\end{lemma} 

The following simple lemma becomes extremely useful when attempting to derive an explicit closed-form formula for some parameters of Algorithm \ref{algorithm.2}.
\begin{lemma}\label{lemma.3} Let $\alpha, \beta, N \in \mathbb{R}_{>0}$. The inequality $N \geq \alpha \log(\beta N)$ holds if $N \geq\max \{\alpha, 2\alpha \log(\alpha\beta)\}$.
\end{lemma}

\subsection{Proof of Proposition \ref{proposition.1}} 
The empirical Bellman operators  $ \{\widehat{\T}_n\}_{n=1}^{N_e}$ are $\gamma$-contractive. Hence, by employing Lemma \ref{lemma.1} and setting $\Theta_H=\Theta^\star$, we have 
\begin{equation*}
   \mathbb{E} \Big\|\frac{1}{N_e}\sum_{n=1}^{N_e}Y_{n+1} -\Theta^\star\Big\|_{\infty}\leq  \frac{2\|\Theta_0-\Theta^\star\|_{\infty}}{(1-\gamma)\lambda N_e} + \frac{1}{1-\gamma}\left(\frac{1-\lambda}{N_e}\sum_{n=1}^{N_e} \mathbb{E}\|P_n^y \|_\infty+ \frac{\lambda}{N_e}\sum_{n=1}^{N_e} \mathbb{E}\|P_n^z \|_\infty\right).
\end{equation*}
Substituting \eqref{Eq:Lemma2.a.1} from Lemma \ref{lemma.2} into the above inequality gives \eqref{Eq:Theorem1.a}. It is worth noting that by using Lemma \ref{lemma.1} and \eqref{Eq:Lemma2.a.2}, one can also derive a high probability bound on the $\ell_\infty$-error.
\subsection{Proof of Theorem \ref{theorem.1}}
The operator $\widehat{H}_n(\Theta)=\widehat{\T}_n (\Theta)-\widehat{\T}_n(\Theta_m)+ \widetilde{\T} (\Theta_m)$ is $\gamma$-contractive. Applying Lemma \ref{lemma.1} with $\Theta_H=\Theta^\star$, and taking into account that $Z_0=Y_0=\Theta_m$, we find that
\begin{equation}\label{Eq:Proof.Theorem2.a}
    \Big\|\underbrace{\frac{1}{N_e} \sum_{n=1}^{N_e} Y_{n+1}}_{\Theta_{m+1}} -\Theta^\star \Big\|_{\infty} \leq  \frac{2\|\Theta_m-\Theta^\star\|_{\infty}}{(1-\gamma)\lambda N_e} + \frac{1}{1-\gamma}\left(\frac{1-\lambda}{N_e}\sum_{n=1}^{N_e} \|P_n^y \|_{\infty}+ \frac{\lambda}{N_e}\sum_{n=1}^{N_e} \|P_n^z \|_{\infty}\right).
\end{equation}

\textbf{Proof of Theorem \ref{theorem.1}.a.} By employing \eqref{Eq:Lemma2.b.1} along with the inequality \eqref{Eq:Proof.Theorem2.a}, we have
 \begin{align*}
     \mathbb{E}\|\Theta_{m+1} -\Theta^\star \|_{\infty} \leq & \left(\underbrace{\frac{2}{(1-\gamma)\lambda N_e}+\frac{2\gamma}{1-\gamma} C(\lambda^2,1)}_{\Psi_{N_e}}+\underbrace{\frac{\gamma}{1-\gamma}C(\frac{1}{N_\T},1)}_{\Psi_{N_\T}}\right) \|\Theta_m-\Theta^\star\|_{\infty}\\
     &+\underbrace {\frac{\gamma}{1-\gamma} C(\frac{1}{N_\T},1)}_{\Psi_{N_\T}} \|\Theta^\star\|_\infty+\underbrace{\frac{1}{1-\gamma}\sqrt{\frac{2\log(2D)}{N_\T}} }_{\Psi_{r}} \sigma_r.
\end{align*}
 A simple calculation shows that choosing the algorithm parameters according to Theorem \ref{theorem.1}.a leads to $\Psi_{N_\T} \leq \frac{\phi^{m+1}}{3}$, $\Psi_{N_e}\leq \frac{\phi}{3}(2-\phi^m)$, and $\Psi_{r} \leq \frac{\phi^{m+1}}{3}$. As a result, we find that \begin{equation}\label{Eq: Proof. Theorem2.inductive}
  \mathbb{E}\|\Theta_{m+1}-\Theta^\star\|_{\infty} \leq \frac{2\phi}{3} \|\Theta_{m}-\Theta^\star\|_{\infty}+ \frac{\phi^{m+1}}{3}  \big( \|\Theta^\star\|_{\infty}+\sigma_r \big).
\end{equation}

Using the above inequality, we prove that $\mathbb{E}\|\Theta_{M}-\Theta^\star\|_\infty \leq \phi^M (\|\Theta^\star\|_\infty+\sigma_r)$ via an inductive argument. The base case is trivial. Now, suppose that $\Theta_m$ satisfies the bound $\mathbb{E}\|\Theta_{m}-\Theta^\star\|_{\infty}\leq \phi^m (\|\Theta^\star\|_\infty+\sigma_r)$. By substituting this inequality into \eqref{Eq: Proof. Theorem2.inductive}, we have $\mathbb{E}\|\Theta_{m+1}-\Theta^\star\|_{\infty}\leq \frac{2}{3}\phi^{m+1} (\|\Theta^\star\|_\infty+\sigma_r)+ \frac{\phi^{m+1}}{3} (\|\Theta^\star\|_\infty+\sigma_r) \leq \phi^{m+1} (\|\Theta^\star\|_\infty +\sigma_r),$ as claimed.

\textbf{Proof of Theorem \ref{theorem.1}.b.} By substituting \eqref{Eq:Lemma2.b.1} into \eqref{Eq:Proof.Theorem2.a}, we find that
 \begin{align*}
     \|\Theta_{m+1}-\Theta^\star \|_{\infty} \leq&\left(\underbrace{\frac{2}{(1-\gamma)\lambda N_e}+\frac{2\gamma}{1-\gamma} C(\lambda^2,\frac{\delta}{5M N_e})}_{\Bar{\Psi}_{N_e}}+\underbrace{\frac{\gamma}{1-\gamma}C(\frac{1}{N_\T},\frac{\delta}{5M})}_{\Bar{\Psi}_{N_\T}}\right) \|\Theta_m-\Theta^\star\|_{\infty}\\
     &+\underbrace {\frac{\gamma}{1-\gamma} C(\frac{1}{N_\T},\frac{\delta}{5M})}_{\Bar{\Psi}_{N_\T}} \|\Theta^\star\|_{\infty}+\underbrace{\frac{1}{1-\gamma}\sqrt{\frac{2\log(\frac{10MD}{\delta})}{N_\T}}}_{\Bar{\Psi}_{r}} \sigma_r,
\end{align*}
 with probability at least $1-\frac{\delta}{M}$. Selecting the algorithm parameters according to Theorem~\ref{theorem.1}.b gives $\Bar{\Psi}_{N_\T} \leq \frac{\phi^{m+1}}{3}$, $\Bar{\Psi}_{r} \leq \frac{\phi^{m+1}}{3}$, and $\Bar{\Psi}_{N_e} \leq \frac{\phi}{3}(2-\phi^m) $, where the last inequality is obtained by applying Lemma \ref{lemma.3}. Consequently, we have $  \|\Theta_{m+1}-\Theta^\star\|_{\infty} \leq \frac{2\phi}{3} \|\Theta_{m}-\Theta^\star\|_{\infty}+ \frac{\phi^{m+1}}{3} \big(\|\Theta^\star\|_{\infty}+\sigma_r\big)$, 
with probability at least $1-\frac{\delta}{M}$. The remainder of the proof relies on an inductive argument identical to that used in the last part of the proof of Theorem \ref{theorem.1}.a, along with the union bound.
\subsection{Proof of Proposition \ref{proposition.2}} \label{Sec:Proof.Proposition 2}
Following a similar analysis as in~\cite[Sec. 4.3]{VRQ.Wainwright}, we show that under the stated conditions in Proposition \ref{proposition.2}, the iterates $\{{\Theta_m}\}_{m=0}^{M}$ satisfy
\begin{equation}\label{Eq:Proof.Proposition.1}
\|\Theta_m-\Theta^\star\|_{\infty} \leq \Bar{c} \phi^m \frac{r_{\text{max}}}{\sqrt{1-\gamma}}.
\end{equation}
with high probability. Note that the above inequality immediately implies $\|\Theta_M-\Theta^\star\|_{\infty}\leq \epsilon$ for $M=\log_\frac{1}{\phi} (\frac{\Bar{c} r_{\text{max}}}{\sqrt{(1-\gamma)}\epsilon})$. We prove \eqref{Eq:Proof.Proposition.1} via an inductive argument. for $m=0$ is trivial. Now, suppose $\|\Theta_m-\Theta^\star\|_{\infty} \leq \Bar{c}\phi^m \frac{r_{\text{max}}}{\sqrt{1-\gamma}}:=b_m$, we show that $\|\Theta_{m+1}-\Theta^\star\|_{\infty}\leq \phi b_m=b_{m+1} $ with probability at least $1-\frac{\delta}{M}$. First, applying Lemma \ref{lemma.1} with $\Theta_H=\widehat{\Theta}$, and taking into account that $Z_0=Y_0=\Theta_m$, we have
\begin{equation}\label{Eq:Proof.proposition.3}
    \|\Theta_{m+1} -\widehat{\Theta} \|_{\infty} \leq  \frac{2\|\Theta_m-\widehat{\Theta}\|_{\infty}}{(1-\gamma)\lambda N_e} + \frac{1}{1-\gamma}\left(\frac{1-\lambda}{N_e}\sum_{n=1}^{N_e} \|P_n^y \|_{\infty}+ \frac{\lambda}{N_e}\sum_{n=1}^{N_e} \|P_n^z \|_{\infty}\right).
\end{equation}
Substituting \eqref{Eq: Lemma2.c.2} into \eqref{Eq:Proof.proposition.3} and using the triangle inequality gives
\begin{equation*}\label{Eq:Proof.Theorem4.c}
    \|\Theta_{m+1} -\widehat{\Theta} \|_{\infty} \leq \Big(\frac{2}{(1-\gamma)\lambda N_e}+\frac{2\gamma}{1-\gamma} C(\lambda^2,1)\Big) \|\Theta_m-\widehat{\Theta}\|_{\infty} \leq \frac{2-\phi^m}{3}b_{m+1}+\frac{\phi(2-\phi^m)}{3}\|\Theta^\star-\widehat{\Theta}\|_{\infty},
\end{equation*}
with probability at least $1-\frac{2\delta}{5M}$. Moreover, according to Lemma 4 in \cite{VRQ.Wainwright}, we have
\begin{align*}
\|\widehat{\Theta}-\Theta^\star\|_{\infty} & \leq \frac{4}{3} b_{m+1} \Big( \frac{\gamma}{\phi(1-\gamma)} C(\frac{1}{N_\T},\frac{\delta}{5 M})  + \frac{ 2 \log(\frac{10 M D}{\delta})} {3\Bar{c} \phi^{m+1} (1-\gamma)^{1.5} N_\T} +\frac{(1+\frac{2\log(2)}{r_{\text{max}}})\sqrt{2\log(\frac{10 M D}{\delta})} }{\Bar{c} \phi^{m+1} (1-\gamma)\sqrt{N_\T}}\Big)\nonumber \\ 
& \leq b_{m+1}\Big(\frac{13}{36} \phi^m+ \frac{\phi^{m+1}}{144} +\frac{1}{48}  \Big),
\end{align*}
with probability at least $1-\frac{3\delta}{5M}$. Finally, by employing the triangle inequality and the union bound, we have
\small
\begin{align*}
    \|\Theta_{m+1}-\Theta^\star\|_{\infty} \leq \|\Theta_{m+1}-\widehat{\Theta}\|_{\infty} +\|\widehat{\Theta}-\Theta^\star\|_{\infty} \leq b_{m+1} \underbrace{\Big( \frac{2-\phi^m}{3}+(\frac{\phi(2-\phi^m)}{3} +1)\big(\frac{13}{36} \phi^m+ \frac{\phi^{m+1}}{144} +\frac{1}{48}  \big) \Big)}_{ \leq 1},
\end{align*}
\normalsize
with probability at least $1-\frac{\delta}{M}$, as claimed.

\subsection{Proof of Theorem \ref{theorem.3}}

Applying Lemma \ref{lemma.1} with $\Theta_H=\widehat{\Theta}$, and taking into account that $Z_0=Y_0=\Theta_m$, we find that
\begin{equation*}
   \mathbb{E}\|\Theta_{m+1} -\widehat{\Theta} \|_{\infty} \leq  \frac{2\mathbb{E} \|\Theta_m-\widehat{\Theta}\|_{\infty}}{(1-\gamma)\lambda N_e} + \frac{1}{1-\gamma}\left(\frac{1-\lambda}{N_e}\sum_{n=1}^{N_e} \mathbb{E}\|P_n^y \|_{\infty}+ \frac{\lambda}{N_e}\sum_{n=1}^{N_e} \mathbb{E}\|P_n^z \|_{\infty}\right).
\end{equation*}
By substituting \eqref{Eq: Lemma2.c.1} into the above inequality and applying the triangle inequality, we find that
\begin{equation} \label{Eq:Proof.Theorem.inc.c}
    \mathbb{E}\|\Theta_{m+1} -\widehat{\Theta} \|_{\infty} \leq \Big(\frac{2}{(1-\gamma)\lambda N_e}+\frac{2\gamma}{1-\gamma} C(\lambda^2,1)\Big)\Big( \mathbb{E} \|\Theta_m-\Theta^\star (\mathcal{P}) \|_{\infty}+\mathbb{E} \|\Theta^\star (\mathcal{P})-\widehat{\Theta}\|_{\infty}\Big).
\end{equation}
Moreover, according to Lemma 4.9 in \cite{Khamaru.TD}, we have
\small
\begin{equation} \label{Eq:Lemma 4.9}
      \mathbb{E} \| \widehat{\Theta}-\Theta^\star (\mathcal{P})\|_{\infty} \leq C(\frac{1}{N_\T},1)\|\Theta_m-\Theta^\star (\mathcal{P})\|_{\infty}   + \Big( \rho(\mathcal{P})+\gamma v(\mathcal{P}) \Big) \sqrt{\frac{2\log 2D}{N_\T}}+ \frac{\|\Theta^\star(\mathcal{P})\|_{\text{span}}} {3 (1-\gamma) N_\T} \log(2 D).
\end{equation}
\normalsize
From the triangle inequality, \eqref{Eq:Proof.Theorem.inc.c}, and \eqref{Eq:Lemma 4.9}, we deduce that
\small
\begin{align*}
 \mathbb{E}\|\Theta_{m+1}&-\Theta^\star (\mathcal{P})\|_{\infty} \leq \mathbb{E}\|\Theta_{m+1}-\widehat{\Theta} \|_{\infty} +\mathbb{E}\|\widehat{\Theta}-\Theta^\star(\mathcal{P})\|_{\infty} \leq  \Big( \underbrace{\frac{2}{(1-\gamma)\lambda N_e}+\frac{2\gamma}{1-\gamma} C(\lambda^2,1)}_{\Psi_{N_e}}+\underbrace{\frac{\gamma}{1-\gamma}C(\frac{1}{N_\T},1)}_{\Psi_{N_\T}}\\
 &+\big(\underbrace{\frac{2}{(1-\gamma)\lambda N_e}+\frac{2\gamma}{1-\gamma} C(\lambda^2,1)}_{\Psi_{N_e}}\big) \underbrace{\frac{\gamma}{1-\gamma}C(\frac{1}{N_\T},1)}_{\Psi_{N_\T}}  \Big)  \|\Theta_{m}-\Theta^\star(\mathcal{P})\|_{\infty}+\Big( \underbrace{\frac{2}{(1-\gamma)\lambda N_e}+\frac{2\gamma}{1-\gamma} C(\lambda^2,1)}_{\Psi_{N_e}}\\
 &+1\Big) \times\Big( \big( \rho(\mathcal{P})+\gamma v(\mathcal{P}) \big) \sqrt{\frac{2\log 2D}{N_\T}}+  \frac{2\|\Theta^\star(\mathcal{P})\|_{\text{span}}} {3 (1-\gamma) N_\T} \log(2 D)   \Big).
\end{align*}
\normalsize
Selecting algorithm parameters according to Theorem \ref{theorem.1}.a results in $\Psi_{N_\T} \leq \frac{\phi^{m+1}}{3}$, and $\Psi_{N_e}\leq \frac{\phi}{3}(2-\phi^m)$. as a result, we have
\small
\begin{align*}
 \mathbb{E}\|\Theta_{m+1}-\Theta^\star(\mathcal{P})\|_{\infty} \leq \Big( \frac{2\phi}{3}+ \frac{\phi^2}{9} \Big)  \|\Theta_{m}-\Theta^\star(\mathcal{P})\|_{\infty}+2 \Big( \big( \rho(\mathcal{P})+\gamma v(\mathcal{P}) \big) \sqrt{\frac{2\log 2D}{N_\T}}+ \frac{\|\Theta^\star\|_{\text{span}}} {3 (1-\gamma) N_\T} \log(2 D) \Big).
\end{align*}
\normalsize
As a direct consequence of the above inequality, we have
\begin{align*}
     \mathbb{E}\|\Theta_{M}-\Theta^\star\|_{\infty} \leq&  \Big( \frac{2\phi}{3}+ \frac{\phi^2}{9} \Big)^M \| \Theta_{0}-\Theta^\star \|_{\infty}+ \sum_{m=0}^{M-1} 2\Big(\frac{2\phi}{3}+ \frac{\phi^2}{9} \big)^{M-1-m}  \Big(\big( \rho(\mathcal{P})+\gamma v(\mathcal{P}) \big) \sqrt{\frac{2\log 2D}{N_\T(m)}}\\ &+\frac{\|\Theta^\star(\mathcal{P})\|_{\text{span}} \log(2 D) } {3 (1-\gamma) N_\T(m) }   \Big).
\end{align*}
Note that $\frac{\phi^{M-1-m}}{\sqrt{N_\T(m)}} =\frac{1}{\sqrt{N_\T(M-1)}}$, and $\frac{\phi^{2(M-1-m)}}{N_\T(m)} =\frac{1}{N_\T(M-1)}$. Hence, we have 
\small
\begin{align}
     \mathbb{E}\|\Theta_{M}-\Theta^\star\|_{\infty} \leq \Big( \frac{2\phi}{3}+ \frac{\phi^2}{9} \Big)^M \| \Theta_{0}-\Theta^\star \|_{\infty} + 9  \big( \rho\mathcal{P})+\gamma v(\mathcal{P}) \big) \sqrt{\frac{2\log 2D}{N_\T(M-1)}}+\frac{1}{\frac{4}{3}-\frac{1}{\phi}} \frac{\|\Theta^\star\|_{\text{span}} \log(2 D) } { (1-\gamma) N_\T(M-1) }.  \label {Eq:Proof.Theorem4.g}
\end{align}
\normalsize
It remains to express the quantities $\Big( \frac{2\phi}{3}+ \frac{\phi^2}{9} \Big)^M$, and $N_\T(M-1)$ in terms of the total number of available samples $N$, and show that the total number of used samples is bounded by $N$. We have
\begin{equation}\label{Eq:Theorem3.equal}
     \Big( \frac{2\phi}{3}+ \frac{\phi^2}{9} \Big)^M=\Big((\frac{1}{\phi})^{-M}\Big)^{1+\log_{\frac{1}{\phi}} (\frac{9}{6+\phi}) }=\Big(\frac {8\sqrt{ \gamma\ \log 2D}} {\sqrt{N}\sqrt{1-\phi^2} (1-\gamma)}\Big)^{1+\log_{\frac{1}{\phi}} (\frac{9}{6+\phi}) }.
\end{equation}
Also, the number of samples used for recentering is 
\begin{equation}\label{Eq:Theorem3.equal.2}
  \sum_{m=0}^{M-1} N_\T(m) \leq \frac{N_\T(M-1)}{(1-\phi^2)} = \frac{32\gamma \log(2D)}{(1-\phi^2)\phi^{2M}(1-\gamma)^2} = \frac{N}{2},  
\end{equation}
where the last equality is obtained via substituting $M=\log_{\frac{1}{\phi}}({\frac{\sqrt{N}\sqrt{1-\phi^2} (1-\gamma)} {8\sqrt{\gamma \log (2D)}} })$.
Substituting \eqref{Eq:Theorem3.equal} and \eqref{Eq:Theorem3.equal.2} into \eqref{Eq:Proof.Theorem4.g} gives \eqref{Eq: Theorem4.Upper bound}. In addition, the number of samples used as epoch lengths is
\begin{equation*}
   \sum_{m=0}^{M-1} N_e(m) \leq M N_e(0)= \log_{\frac{1}{\phi}}( \frac {\sqrt{N}\sqrt{1-\phi^2}(1-\gamma)} {8\sqrt{\gamma\ \log( 2D)}}) N_e(0)   \leq \frac{N}{2},
\end{equation*}
where the second inequality is an immediate consequence of inequality \eqref{Eq: Theorem4.sample size} and Lemma \ref{lemma.3}. As a result, the total number of samples used by VRCQ is 
\begin{equation*}
    \sum_{m=0}^{M-1} N_e(m) + \sum_{m=0}^{M-1} N_\T(m) \leq \frac{N}{2}+ \frac{N}{2} =N.
\end{equation*}

\begin{appendices}
\section{Proofs of the Auxiliary Lemmas}
In this appendix, we provide the proofs of the auxiliary lemmas used in Section~\ref{sec: proofs}.

\subsection{Proof of Lemma \ref{lemma.1}}
We begin by presenting a "sandwich result" that provides both lower and upper bounds for the error sequences \(\{Y_n - \Theta_H\}_{n=1}^{N_e}\) and \(\{Z_n - \Theta_H\}_{n=1}^{N_e}\).

\begin{lemma}\label{lemma.4}
The sequences $\{Y_{n}\}_{n=1}^{N_e}$ and $\{Z_{n}\}_{n=1}^{N_e}$ generated by the recursion \eqref{Eq:Lemma1.Y} and \eqref{Eq:Lemma1.Z} satisfy the following sandwich inequality
\begin{equation}\label{Eq:Sandwich}
    \left[ \begin{array}{c}
      P_{n}^y-a_{n}^y \mathds{1} \\
      P_{n}^z-a_{n}^z \mathds{1}
    \end{array}\right] \leq  \left[ \begin{array}{c}
      Y_n -\Theta_H \\
      Z_n -\Theta_H 
    \end{array}\right] \leq \left[ \begin{array}{c}
      a_{n}^y \mathds{1}+P_{n}^y \\
      a_{n}^z \mathds{1}+P_{n}^z 
    \end{array}\right], 
\end{equation}
where the non-negative scalar sequences $a_n^y$, $a_n^z$ are generated by the linear dynamics 
\begin{align*}
    &\left[ \begin{array}{c}
      a_{n+1}^y \\
      a_{n+1}^z 
    \end{array}\right]=\underbrace{\left[ \begin{array}{cc}
       1-\lambda  & \lambda \\
        (1-\lambda)\lambda\gamma   & 1-\lambda+\lambda^2 \gamma
    \end{array}\right]}_{H} \left[ \begin{array}{c}
      a_{n}^y \\
      a_{n}^z 
    \end{array}\right] +\underbrace{\left[ \begin{array}{cc}
      0  & 0 \\
       (1-\lambda)\lambda\gamma   &  \lambda^2 \gamma  
    \end{array}\right]}_{F}\left[ \begin{array}{c}
      \|P_n^y \|_{\infty} \\
      \|P_n^z\|_{\infty} 
    \end{array}\right],
\end{align*}
with initial conditions $a_1^y=\|Y_1-\Theta^\star\|_{\infty}$ and $a_1^z=\|Z_1-\Theta^\star\|_{\infty}$.\par 
\end{lemma}

By employing Lemma \ref{lemma.4}, we have
\begin{equation*}
     \frac{1}{N_e}\sum_{n=1}^{N_e} \left[ \begin{array}{c}
      P_{n}^y-a_{n}^y \mathds{1} \\
      P_{n}^z-a_{n}^z \mathds{1} 
    \end{array}\right] \leq \frac{1}{N_e}\sum_{n=1}^{N_e} \left[ \begin{array}{c}
      Y_n -\Theta_H \\
      Z_n -\Theta_H 
    \end{array}\right] \leq  \frac{1}{N_e}\sum_{n=1}^{N_e} \left[ \begin{array}{c}
      P_{n}^y+a_{n}^y \mathds{1} \\
      P_{n}^z+a_{n}^z \mathds{1} 
    \end{array}\right],
\end{equation*}
which implies 
\begin{equation*}
        \left[ \begin{array}{c}
      \|\frac{1}{N_e}\sum_{n=1}^{N_e}\left(Y_n -\Theta_H-P_{n}^y\right)\|_{\infty} \\
      \|\frac{1}{N_e}\sum_{n=1}^{N_e}\left(Z_n -\Theta_H-P_{n}^z\right)\|_{\infty} 
    \end{array}\right] \leq  \frac{1}{N_e}\sum_{n=1}^{N_e} \left[ \begin{array}{c}
      a_{n}^y \\
      a_{n}^z 
    \end{array}\right].
\end{equation*}
Applying the triangle inequality leads to
\begin{equation}\label{eq:Proof.Lemma1.a}
     \left[ \begin{array}{c}
      \|\frac{1}{N_e}\sum_{n=1}^{N_e} Y_n -\Theta_H\|_{\infty} \\
      \|\frac{1}{N_e}\sum_{n=1}^{N_e}Z_n -\Theta_H\|_{\infty} 
    \end{array}\right] \leq  \frac{1}{N_e}\sum_{n=1}^{N_e} \left[ \begin{array}{c}
      a_{n}^y \\
      a_{n}^z 
    \end{array}\right]+\frac{1}{N_e}\sum_{n=1}^{N_e} \left[ \begin{array}{c}
     \|P_{n}^y\|_{\infty} \\
      \|P_{n}^z\|_{\infty} 
    \end{array}\right].
\end{equation}

The lemma below establishes an upper bound on $\sum_{n=1}^{N_e} \left[ \begin{array}{c}
      a_{n}^y \\
      a_{n}^z 
    \end{array}\right]$.
    
\begin{lemma}\label{lemma.5}
The sequences $\{a_n^y\}_{n=1}^{N_e}$ and $\{a_n^z\}_{n=1}^{N_e}$ satisfy the following inequality 

 \begin{align} \label{Eq:Lemma.5.main}
    \sum_{n=1}^{N_e} \left[ \begin{array}{c}
      a_{n}^y \\
      a_{n}^z 
    \end{array}\right] \leq (1-H)^{-1} \left( \left[ \begin{array}{c}
      a_{1}^y \\
      a_{1}^z 
    \end{array}\right] +F \sum_{n=1}^{N_e} \left[ \begin{array}{c}
      \|P_n^y \|_{\infty} \\
      \|P_n^z\|_{\infty}
      \end{array}\right] \right).
 \end{align}
\end{lemma}
By utilizing Lemma \ref{lemma.5} and taking into account the fact that $a_1^z=a_1^y=\|\Theta_0-\Theta^\star\|_{\infty}$,  we find that
\begin{align*}
(1-H)^{-1} \left[ \begin{array}{c}
      a_{1}^y \\
      a_{1}^z 
    \end{array}\right]=\frac{\|\Theta_0-\Theta^\star\|_{\infty}}{(1-\gamma)\lambda}\left[ \begin{array}{c}
        2-\lambda\gamma \\
       (1-\lambda)\gamma+1
      \end{array}\right] \leq \frac{2\|\Theta_0-\Theta^\star\|_{\infty}}{(1-\gamma)\lambda}\left[ \begin{array}{c}
       1 \\
       1
      \end{array}\right].    
\end{align*}
Moreover, we have
\begin{equation*}
    (1-H)^{-1} F \sum_{n=1}^{N_e} \left[ \begin{array}{c}
      \|P_n^y \|_{\infty} \\
      \|P_n^z\|_{\infty}
      \end{array}\right]=\frac{\gamma}{1-\gamma}\left((1-\lambda)\sum_{n=1}^{N_e} \|P_n^y \|_{\infty}+\lambda \sum_{n=1}^{N_e} \|P_n^z \|_{\infty}\right)\left[ \begin{array}{c}
       1 \\
       1
      \end{array}\right].
\end{equation*}
Therefore, according to \eqref{Eq:Lemma.5.main}, we can conclude that
\begin{equation}\label{eq:proof.Lemma1.an}
      \frac{1}{N_e}\sum_{n=1}^{N_e} \left[ \begin{array}{c}
      a_{n}^y \\
      a_{n}^z 
    \end{array}\right] \leq\left( \frac{2\|\Theta_0-\Theta^\star\|_{\infty}}{(1-\gamma)\lambda N_e}+\frac{\gamma}{1-\gamma}\Big(\frac{1-\lambda}{N_e}\sum_{n=1}^{N_e} \|P_n^y \|_{\infty}+ \frac{\lambda}{N_e}\sum_{n=1}^{N_e} \|P_n^z \|_{\infty}\Big)\right)\left[ \begin{array}{c}
       1 \\
       1
      \end{array}\right].
\end{equation}
Finally, by utilizing \eqref{eq:Proof.Lemma1.a}, \eqref{eq:proof.Lemma1.an} and the triangle inequality, we find that
\begin{align*}
    \|\frac{1}{N_e}\sum_{n=1}^{N_e}Y_{n+1} -\Theta^\star\|_{\infty} &\leq (1-\lambda) \|\frac{1}{N_e}\sum_{n=1}^{N_e}Y_n -\Theta^\star\|_{\infty}+\lambda\|\frac{1}{N_e}\sum_{n=1}^{N_e}Z_n -\Theta^\star\|_{\infty} \\
    &\leq \sum_{n=1}^{N_e} \left( \frac{1-\lambda}{N_e} (a_{n}^y+ \|P_n^y\|_{\infty})+ \frac{\lambda}{N_e} (a_{n}^z+ \|P_n^z\|_{\infty})\right)\\
    &\leq  \frac{2\|\Theta_0-\Theta_H\|_{\infty}}{(1-\gamma)\lambda N_e} + \frac{1}{1-\gamma}\left(\frac{1-\lambda}{N_e}\sum_{n=1}^{N_e} \|P_n^y \|_{\infty}+ \frac{\lambda}{N_e}\sum_{n=1}^{N_e} \|P_n^z \|_{\infty}\right) ,
\end{align*}
as claimed.
\subsection{Proof of Lemma \ref{lemma.2}}

\subsubsection{Proof of Lemma \ref{lemma.2}.a}
$W_n(x,u)$ can be written as $W_n(x,u)=\Bar{W}_n (x,u) + \widehat{W}_n (x,u)$, where $\Bar{W}_n:=\hat{r}(x,u)-r(x,u)$ and $\widehat{W}_n (x,u):=\gamma \Big( \max_{\Bar{u}} \Theta^\star(x_n,\Bar{u})- \mathbb{E}_{\Bar{x}} \max_{\Bar{u}} \Theta^\star(\Bar{x},\Bar{u})\Big)$. Since $\Bar{W}_n(x,u)$ is $\sigma_r$-sub-Gaussian, we have $\log ( \mathbb{E}[ e^{s \Bar{W}_n(x,u)} ]) \leq \frac{s^2 \sigma_r^2}{2}$. Moreover, $\widehat{W}_n(x,u)$ is bounded in absolute value by  $\gamma \|\Theta^\star\|_{\text{span}}$ and has variance $\sigma^2(\Theta^\star)(x,u)$. Hence, it satisfies Bernstein's condition (see, e.g.,~\cite{Bern.B,Bern.W})
\begin{equation*}
    \log ( \mathbb{E} e^{s \widehat{W}(x,u)} )\leq \frac{\frac{1}{2}s^2 \sigma^2(\Theta^\star)(x,u) }{1-\frac{1}{3}|s| \gamma \|\Theta^\star\|_{\text{span}}},\;\;\; \forall|s|\leq \frac{3}{\gamma\|\Theta^\star\|_{\text{span}}}.
\end{equation*}
The processes $\{P_n^z\}_{n=1}^{N_e}$ and $\{P_n^y\}_{n=1}^{N_e}$ can be expressed as $P_n^z= \Bar{P}_n^z+\widehat{P}_n^z $
and $P_n^y= \Bar{P}_n^y+\widehat{P}_n^y $, where
\begin{align*}
    \Bar{P}_{n+1}^z=(1-\lambda) \Bar{P}_{n}^z + \lambda \Bar{W}_n,\;\;\;    \widehat{P}_{n+1}^z=(1-\lambda) \widehat{P}_{n}^z + \lambda \widehat{W}_n  \\
    \Bar{P}_{n+1}^y=(1-\lambda) \Bar{P}_{n}^y + \lambda \Bar{P}_{n}^z,\;\;\;  \widehat{P}_{n+1}^y=(1-\lambda) \widehat{P}_{n}^y + \lambda \widehat{P}_{n}^z 
\end{align*}
and $\Bar{P}_1^z=\widehat{P}_1^z=\Bar{P}_1^y=\widehat{P}_1^y=0$. We now provide bounds on the moment generating functions of the stochastic processes $\Bar{P}_n^z$, $\widehat{P}_n^z$, $\Bar{P}_n^y$, and $\widehat{P}_n^y$ through an inductive argument. For $n=1$ we have $ \log( \mathbb{E} e^{s \Bar{P}_{1}^z(x,u)})= \log( \mathbb{E} e^{s \widehat{P}_{1}^z(x,u)})=  \log( \mathbb{E} e^{s \Bar{P}_{1}^y(x,u)})=\log( \mathbb{E} e^{s \widehat{P}_{1}^y(x,u)})=0$.
Assume that
\begin{subequations}
\begin{align}
   \log(\mathbb{E} e^{s \Bar{P}_{n}^z(x,u)}) \leq \frac{\lambda \sigma_r^2s^2}{2}, \;\;  \log( \mathbb{E} e^{s \widehat{P}_{n}^z(x,u)}) \leq  \frac{\frac{1}{2}s^2 \lambda \sigma^2(\Theta^\star)(x,u) }{1-\frac{1}{3}|s|\lambda\gamma \|\Theta^\star\|_{\text{span}}}, \label{Eq:proof.lemma2.ind.1}\\
    \log( \mathbb{E} e^{s \Bar{P}_{n}^y(x,u)}) \leq \frac{\lambda^2 \sigma_r^2s^2}{2}, \;\;  \log( \mathbb{E} e^{s \widehat{P}_{n}^y(x,u)}) \leq  \frac{\frac{1}{2}s^2 \lambda^2 \sigma^2(\Theta^\star)(x,u) }{1-\frac{1}{3}|s|\lambda^2\gamma \|\Theta^\star\|_{\text{span}}},  \label{Eq:proof.lemma2.ind.2}
\end{align}
\end{subequations}
then we have 
\begin{align*}
    &\log (\mathbb{E} e^{s \Bar{P}_{n+1}^z(x,u)}) \leq (1-\lambda) \frac{\lambda \sigma_r^2s^2}{2} + \lambda  \frac{\lambda \sigma_r^2s^2}{2} =\frac{\lambda \sigma_r^2s^2}{2},\\
    &\log (\mathbb{E} e^{s \Bar{P}_{n+1}^y(x,u)}) \leq (1-\lambda) \frac{\lambda^2 \sigma_r^2s^2}{2} + \lambda  \frac{\lambda^2 \sigma_r^2s^2}{2} =\frac{\lambda^2 \sigma_r^2s^2}{2}\\
    &\log (\mathbb{E} e^{s \widehat{P}_{n+1}^z(x,u)}) \leq (1-\lambda) \frac{\frac{1}{2}s^2 \lambda \sigma^2(\Theta^\star)(x,u) }{1-\frac{1}{3}|s|\lambda \gamma \|\Theta^\star\|_{\text{span}}} + \lambda  \frac{\frac{1}{2}s^2 \lambda \sigma^2(\Theta^\star)(x,u) }{1-\frac{1}{3}|s|\lambda\gamma \|\Theta^\star\|_{\text{span}}} = \frac{\frac{1}{2}s^2 \lambda \sigma^2(\Theta^\star)(x,u) }{1-\frac{1}{3}|s|\lambda\gamma \|\Theta^\star\|_{\text{span}}} \\
    &\log (\mathbb{E} e^{s \widehat{P}_{n+1}^y(x,u)}) \leq (1-\lambda) \frac{\frac{1}{2}s^2 \lambda^2 \sigma^2(\Theta^\star)(x,u) }{1-\frac{1}{3}|s|\lambda^2 \gamma \|\Theta^\star\|_{\text{span}}} + \lambda   \frac{\frac{1}{2}s^2 \lambda^2 \sigma^2(\Theta^\star)(x,u) }{1-\frac{1}{3}|s|\lambda^2\gamma \|\Theta^\star\|_{\text{span}}}=  \frac{\frac{1}{2}s^2 \lambda^2 \sigma^2(\Theta^\star)(x,u) }{1-\frac{1}{3}|s|\lambda^2\gamma \|\Theta^\star\|_{\text{span}}}
\end{align*}
\normalsize
Hence, the inequalities \eqref{Eq:proof.lemma2.ind.1} and \eqref{Eq:proof.lemma2.ind.2} hold for all $n = 1, \ldots, N_e$. Moreover, since $e^{s\| \Bar{P}_{n}\|_\infty} \leq \sum_{(x,u)} \big(e^{s \Bar{P}_{n}^z(x,u)}+ e^{-s \Bar{P}_{n}^z(x,u)}\big) $, we find that
\begin{equation*}
   \mathbb{E} e^{s \|\Bar{P}_{n}^z\|_{\infty}} \leq  \sum_{(x,u)} \left( \mathbb{E} e^{s \Bar{P}_{n}^z(x,u)}+\mathbb{E} e^{-s \Bar{P}_{n}^z(x,u)} \right) \leq 2 D e^{\frac{\lambda \sigma_r^2s^2}{2}}. 
\end{equation*}
Similarly, we have
\begin{equation*}
\mathbb{E} e^{s \|\Bar{P}_{n}^y\|_{\infty}}\leq 2 D e^{\frac{\lambda \sigma_r^2s^2}{2}},\;\;   \mathbb{E} e^{s \|\widehat{P}_{n}^z\|_{\infty}} \leq 2D\frac{\frac{1}{2}s^2 \lambda\|\sigma(\Theta^\star)\|_{\infty}^2 }{1-\frac{1}{3}|s|\lambda\gamma \|\Theta^\star\|_{\text{span}}},\;\;  \mathbb{E} e^{s \|\widehat{P}_{n}^y\|_{\infty}} \leq 2D\frac{\frac{1}{2}s^2 \lambda^2\|\sigma(\Theta^\star)\|_{\infty}^2 }{1-\frac{1}{3}|s|\lambda^2\gamma \|\Theta^\star\|_{\text{span}}}    
\end{equation*}

\textbf{Proof of bound \eqref{Eq:Lemma2.a.1}:} By employing the Jensen's inequality, we find that 
 \begin{align*}
 & \mathbb{E} \|\Bar{P}_{n}^z\|_{\infty}\leq \frac{\lambda \sigma^2s}{2}+ \frac{\log(2D)}{s} ,\;\; \mathbb{E} \|\widehat{P}_{n}^z\|_{\infty}\leq \frac{\frac{1}{2}s \lambda \|\sigma(\Theta^\star)\|^2_{\infty} }{1-\frac{1}{3}s\lambda\gamma \|\Theta^\star\|_{\text{span}}}+\frac{\log(2D)}{s} \\
 &\mathbb{E} \|\Bar{P}_{n}^y\|_{\infty}\leq \frac{\lambda^2 \sigma^2s}{2}+ \frac{\log(2D)}{s}   ,\;\; \mathbb{E}\|\widehat{P}_{n}^y\|_{\infty} \leq \frac{\frac{1}{2}s \lambda^2\|\sigma(\Theta^\star)\|^2_{\infty} }{1-\frac{1}{3}s\lambda^2\gamma \|\Theta^\star\|_{\text{span}}}+\frac{\log(2D)}{s}.
 \end{align*}
  Minimizing the right-hand side of the above inequalities with respect to $s$ results in 
  \begin{align*}
 & \mathbb{E} \|\Bar{P}_{n}^z\|_{\infty}\leq  \sqrt{2 \lambda \log(2D)} \sigma_r ,\;\; \mathbb{E} \|\widehat{P}_{n}^z\|_{\infty}\leq \frac{\gamma}{3}\lambda\log(2D) \|\Theta^\star\|_{\text{span}}  + \sqrt{2\lambda \log(2D)} \|\sigma(\Theta^\star)\|_{\infty} \\
 &\mathbb{E} \|\Bar{P}_{n}^y\|_{\infty}\leq \sqrt{2 \lambda^2 \log(2D)} \sigma_r ,\;\; E\mathbb{E} \|\widehat{P}_{n}^y\|_{\infty} \leq \frac{\gamma}{3}\lambda^2\log(2D) \|\Theta^\star\|_{\text{span}} + \sqrt{2 \lambda^2 \log(2D)} \|\sigma(\Theta^\star)\|_{\infty}.
 \end{align*}
 Moreover, using the triangle inequality, we find that
 \begin{align*}
     & \mathbb{E} \|P_n^z\|_{\infty} \leq \mathbb{E} \|\Bar{P}_n^z\|_{\infty}+\mathbb{E}\|\widehat{P}_n^z\|_{\infty}=\frac{\gamma}{3}\lambda\log(2D) \|\Theta^\star\|_{\text{span}}  + \sqrt{2\lambda \log(2D)} \big(\|\sigma(\Theta^\star)\|_\infty + \sigma_r\big), \\
     & \mathbb{E} \|P_n^y\|_{\infty} \leq \mathbb{E} \|\Bar{P}_n^y\|_{\infty}+\mathbb{E} \|\widehat{P}_n^y\|_{\infty}=\frac{\gamma}{3}\lambda^2\log(2D) \|\Theta^\star\|_{\text{span}} + \sqrt{2 \lambda^2 \log(2D)} \big(\|\sigma(\Theta^\star)\|_\infty + \sigma_r\big).
 \end{align*}
 Finally, we have
 \begin{align*}
   \frac{1-\lambda}{N_e}\sum_{n=1}^{N_e} \mathbb{E} \|P_n^y \|_{\infty}+ \frac{\lambda}{N_e}\sum_{n=1}^{N_e} \mathbb{E} \|P_n^z \|_{\infty} &\leq \gamma\underbrace{\Big((1-\lambda)\frac{\lambda^2}{3}+\frac{\lambda^2}{3}\Big)}_{\leq \frac{2}{3}\lambda^2}\log(2D) \|\Theta^\star\|_{\text{span}}\\
   &+\underbrace{\Big((1-\lambda)\lambda+\lambda\sqrt{\lambda}\Big)}_{\leq 2\lambda}\sqrt{2\log(2D)} \big(\|\sigma(\Theta^\star)\|_{\infty} + \sigma_r\big),
\end{align*}
which establishes the claim.

\textbf{Proof of bound \eqref{Eq:Lemma2.a.2}:} Emplying the exponential Chebyshev's inequality leads to
\begin{align*}
 &  \|\Bar{P}_{n}^z\|_{\infty}\leq  \sqrt{2 \lambda \log(\frac{2D}{\delta})} \sigma_r ,\;\; \|\widehat{P}_{n}^z\|_{\infty}\leq \frac{\gamma}{3}\lambda\log(\frac{2D}{\delta}) \|\Theta^\star\|_{\text{span}}  + \sqrt{2\lambda \log(\frac{2D}{\delta})} \|\sigma(\Theta^\star)\|_{\infty} \\
 & \|\Bar{P}_{n}^y\|_{\infty}\leq \sqrt{2 \lambda^2 \log(\frac{2D}{\delta})} \sigma_r ,\;\; \|\widehat{P}_{n}^y\|_{\infty} \leq \frac{\gamma}{3}\lambda^2\log(\frac{2D}{\delta}) \|\Theta^\star\|_{\text{span}} + \sqrt{2 \lambda^2 \log(\frac{2D}{\delta})} \|\sigma(\Theta^\star)\|_{\infty}.
 \end{align*}
with probability at least $1- \delta$. By applying the union bound, we obtain
\begin{align*}
   \frac{1-\lambda}{N_e}\sum_{n=1}^{N_e} \|P_n^y \|_{\infty}+ \frac{\lambda}{N_e}\sum_{n=1}^{N_e} \|P_n^z \|_{\infty} \leq & \frac{2\gamma}{3} \lambda^2 \log(\frac{8DN_e}{\delta}) \|\Theta^\star\|_{\text{span}}+2\lambda \sqrt{2\log(\frac{8DN_e}{\delta})} \big(\|\sigma(\Theta^\star)\|_{\infty} + \sigma_r\big)
\end{align*}
With probability at least $1-\delta$, as claimed.

\subsubsection{Proof of Lemma \ref{lemma.2}.b}
The random matrix $W_n$ can be written as $W_n=\widehat{W}_n+W^\dagger+W^\circ$, where $\widehat{W}_n:=\widehat{\T}_n(\Theta^\star)-\widehat{\T}_n(\Theta_m)+\T(\Theta_m)-\T(\Theta^\star)$, $W^\dagger:=\widetilde{\T}(\Theta_m)-\widetilde{\T}(\Theta^\star)-\T(\Theta_m)+\T(\Theta^\star)$, and $W^\circ:=\widetilde{\T}(\Theta^\star)-\T(\Theta^\star)$. Consequently, the stochastic processes $P_n^z$ and $P_n^y$ can be expressed as $P_n^z=\widehat{P}_n^z+\Bar{P}_n^z$ and $P_n^y=\widehat{P}_n^y+\Bar{P}_n^z$, where 
\begin{align*}
    \widehat{P}^y_{n+1}&= (1-\lambda)\widehat{P}^y_n+\lambda \widehat{P}_n^z, \;\; \Bar{P}^y_{n+1}=(1-\lambda) \Bar{P}_n^z +\lambda \Bar{P}_n^z \\ \label{Eq:recursion.Pzhat}
    \widehat{P}^z_{n+1}&=(1-\lambda)\widehat{P}^z_n+\lambda \widehat{W}_n,    \;\; \Bar{P}^z_{n+1}=(1-\lambda) \Bar{P}_n^z +\lambda (W^\circ+W^\dagger) 
\end{align*}
with $\widehat{P}_1^y=\widehat{P}_1^z=\Bar{P}_1^y=\Bar{P}_1^z=0$. Since $W^\circ$ and $W^\dagger$ are independent of $n$, it follows that
\begin{equation*} 
    \Bar{P}^z_n=\frac{1-(1-\lambda)^{n-1}}{1-(1-\lambda)}\lambda (W^\circ+W^\dagger) \rightarrow \|\Bar{P}^z_n\|_{\infty}\leq \|W^\circ\|_{\infty}+\|W^\dagger\|_{\infty}.
\end{equation*}
Also, we have $\Bar{P}^y_n=\sum_{i=1}^{n-1}(1-\lambda)^{n-1-i}\lambda \Bar{P}^z_i$. As a result, by using the triangle inequality, we have
\begin{equation*}
          \|\Bar{P}^y_n\|_{\infty}\leq \sum_{i=1}^{n-1}(1-\lambda)^{n-1-i}\lambda \|\Bar{P}^z_i\|_{\infty}\leq \frac{1-(1-\lambda)^{n-1}}{1-(1-\lambda)} \lambda (\|W^\circ\|_{\infty}+\|W^\dagger\|_{\infty}) \leq \|W^\circ\|_{\infty}+\|W^\dagger\|_{\infty}.
\end{equation*}
Consequently, we find that
\begin{equation}\label{Eq:Proof.Lemma2.b.1.2}
     \frac{1-\lambda}{N_e}\sum_{n=1}^{N_e} \|P_n^y \|_{\infty}+\frac{\lambda}{N_e}\sum_{n=1}^{N_e} \|P_n^z \|_{\infty}  \leq \frac{1-\lambda}{N_e}\sum_{n=1}^{N_e}\|\widehat{P}_n^y \|_{\infty}+ \frac{\lambda}{N_e}\sum_{n=1}^{N_e} \|\widehat{P}_n^z \|_{\infty}+\|W^\circ\|_{\infty}+\|W^\dagger\|_{\infty}.
\end{equation}
Next, we bound each term on the right-hand side of the above inequality.\par
\textbf{Upper bound on $\|\widehat{P}_n^y \|_{\infty}$ and $\|\widehat{P}_n^z \|_{\infty}$:} The random variable $\widehat{W}(x,u)$ is bounded in absolute value by $2 \gamma \|\Theta_m-\Theta^\star\|_{\infty}$ and its variance is at most $\gamma^2\|\Theta_m-\Theta^\star\|_{\infty}^2$.
As a result, it satisfies Bernstein's condition
\begin{equation*}
    \log ( \mathbb{E} e^{s \widehat{W}(x,u)} ) \leq \frac{\frac{1}{2}s^2\gamma^2 \|\Theta_m-\Theta^\star\|_{\infty}^2 }{1-\frac{2}{3}|s|\gamma\|\Theta_m-\Theta^\star\|_{\infty}},\;\; \forall|s|\leq \frac{3}{2\gamma\|\Theta_m-\Theta^\star\|_{\infty}}.
\end{equation*}

By using an inductive argument similar to the proof of \eqref{Eq:Lemma2.a.1}, we find that
\begin{equation*}
  \log (\mathbb{E} e^{s \widehat{P}_{n}^z(x,u)}) \leq\frac{\frac{1}{2}s^2 \lambda^2\gamma^2 \|\Theta_m-\Theta^\star\|_{\infty}^2 }{1-\frac{2}{3}|s|\lambda\gamma \|\Theta_m-\Theta^\star\|_{\infty}}, \;\;\; \log (\mathbb{E} e^{s \widehat{P}_{n}^y(x,u)}) \leq \frac{\frac{1}{2}s^2 \lambda^2\gamma^2 \|\Theta_m-\Theta^\star\|^2_{\infty} }{1-\frac{2}{3}|s|\lambda^2\gamma \|\Theta_m-\Theta^\star\|_{\infty}}
\end{equation*}
 for $n=1,...,N_e$. Again, using the same lines of arguments as in the proof of Lemma \ref{lemma.2}.a, we have
 \begin{align*}
     &\frac{1-\lambda}{N_e}\sum_{n=1}^{N_e} \mathbb{E} \|\widehat{P}_n^y \|_{\infty}+ \frac{\lambda}{N_e}\sum_{n=1}^{N_e} \mathbb{E} \|\widehat{P}_n^z \|_{\infty} \leq 2 \gamma  C(\lambda^2,1)  \|\Theta_m-\Theta^\star\|_{\infty},\\
     & \frac{1-\lambda}{N_e}\sum_{n=1}^{N_e} \|\widehat{P}_n^y \|_{\infty}+ \frac{\lambda}{N_e}\sum_{n=1}^{N_e} \|\widehat{P}_n^z \|_{\infty} \leq  2 \gamma C(\lambda^2,\frac{\delta}{N_e}) \|\Theta_m-\Theta^\star\|_{\infty}
 \end{align*}
 with probability at least $1-2 \delta$.
 
\textbf{Upper bound on $\| W^\circ \|_{\infty}$:} The random variable $ W^\circ (x,u)$ can be written as 
\begin{equation*}
    W^\circ(x,u) = \underbrace{\sum_{i=1}^{N_e} \frac{1}{N_\T}\Big(\hat{r}_i(x,u)-r(x,u) \Big)}_{:=\Bar{W}^\circ} + \underbrace  {\sum_{i=1}^{N_e} \frac{\gamma }{N_\T} \Big( \max_{\Bar{u}} \Theta^\star(x_i,\Bar{u})- \mathbb{E}_{\Bar{x}} \max_{\Bar{u}} \Theta^\star(\Bar{x},\Bar{u})\Big)}_{:=\widehat{W}^\circ}
\end{equation*}
$\Bar{W}^\circ(x,u)$ is the sum of $N_e$ i.i.d $\sigma_r$-sub Gaussian random variable and $\widehat{W}^\circ(x,u)$ is the sum of $N_\T$ independent and identically distributed random variables, where each of these variables has variance $\sigma^2(\Theta^\star)(x,u) \leq \gamma^2\|\Theta^\star\|_{\infty}^2$ and is bounded in absolute value by $\|\Theta^\star\|_{\text{span}} \leq 2\gamma\|\Theta^\star\|_{\infty}$. Hence, we have $\log \Big(\mathbb{E} e^{ \Bar{W}^\circ (x,u) }\Big) \leq  \frac{\sigma_r^2s^2}{2N_\T}$, and $\log \Big( \mathbb{E} e^{ \widehat{W}^\circ(x,u)} \Big) \leq  \frac{\frac{1}{2N_\T}s^2 \gamma^2 \|\Theta^\star\|_{\infty}^2 }{1-\frac{2}{3N_\T}|s| \gamma\|\Theta^\star\|_{\infty}}$. Moreover, following the same lines of arguments as in the proof of Lemma \ref{lemma.2}.a, we find that 
\begin{align*}
   &\mathbb{E} \| \Bar{W}^\circ\|_{\infty} \leq \sqrt{\frac{2\log(2D)}{N_\T}} \sigma_r  ,\;\;  \mathbb{E} \| \widehat{W}^\circ \|_{\infty} \leq \gamma C(\frac{1}{N_\T},1)\|\Theta^\star\|_{\infty},\\
   &\| \Bar{W}^\circ \|_{\infty} \leq  \sqrt{\frac{2\log(\frac{2D}{\delta})}{N_\T}} \sigma_r  ,\;\;   \| \widehat{W}^\circ \|_{\infty} \leq  \gamma C(\frac{1}{N_\T},\delta) \|\Theta^\star\|_{\infty}
\end{align*}
with probability at least $1-\delta$. By applying the triangle inequality and union bound, we have
\begin{align*}
   & \mathbb{E} \| W^\circ\|_{\infty} \leq   E \| \Bar{W}^\circ\|_{\infty} + \mathbb{E} \| \widehat{W}^\circ \|_{\infty} \leq \gamma C(\frac{1}{N_\T},1)\|\Theta^\star\|_{\infty} + \sqrt{\frac{2\log(2D)}{N_\T}} \sigma_r         ,\\
   & \| W^\circ\|_{\infty} \leq  \| \Bar{W}^\circ\|_{\infty} + \| \widehat{W}^\circ \|_{\infty} \leq \gamma C(\frac{1}{N_\T},\delta)\|\Theta^\star\|_{\infty} + \sqrt{\frac{2\log(\frac{2D}{\delta})}{N_\T}} \sigma_r   
\end{align*}
with probability at least $1-2\delta$.

\textbf{Upper bound on $\|W^\dagger\|_{\infty}$}: The random variable $W^\dagger(x,u)$ is the sum of $N_\T$ i.i.d random variables. Each of these term is bounded in absolute value by $2\gamma\|\Theta_m-\Theta^\star\|_{\infty}$, and has a variance at most $\gamma^2\|\Theta_m-\Theta^\star\|_{\infty}^2$. Therefore, we have $\mathbb{E} \|W^\dagger\|_{\infty} \leq \gamma C(\frac{1}{N_\T},1) \|\Theta_m-\Theta^\star\|_{\infty}$, and $\|W^\dagger\|_{\infty} \leq \gamma C(\frac{1}{N_\T},\delta) \|\Theta_m-\Theta^\star\|_{\infty}$, with probability at least $1-\delta$.

\textbf{Putting together the pieces}: By substituting the above inequalities into \eqref{Eq:Proof.Lemma2.b.1.2} and applying the union bound, we obtain
\begin{align*}
    \frac{1-\lambda}{N_e}\sum_{n=1}^{N_e} \mathbb{E} \|P_n^y \|_{\infty}+ \frac{\lambda}{N_e}\sum_{n=1}^{N_e} \mathbb{E} \|P_n^z \|_{\infty}  \leq& \big(2\gamma C(\lambda^2, 1 ) + \gamma C(\frac{1}{N_\T}, 1)\big) \|\Theta_m-\Theta^\star\|_{\infty}\\
    &+\gamma C(\frac{1}{N_\T},1)\|\Theta^\star\|_{\infty}
    +\sqrt{\frac{2\log(2D)}{N_\T}} \sigma_r   \\
    \frac{1-\lambda}{N_e}\sum_{n=1}^{N_e} \|P_n^y \|_{\infty}+ \frac{\lambda}{N_e}\sum_{n=1}^{N_e} \|P_n^z \|_{\infty}  \leq& \big(2\gamma C(\lambda^2, \frac{\delta}{5 N_e M} ) + \gamma C(\frac{1}{N_\T}, \frac{\delta}{5 M})\big) \|\Theta_m-\Theta^\star\|_{\infty}\\
    &+\gamma C(\frac{1}{N_\T},\frac{\delta}{5M})\|\Theta^\star\|_{\infty} + \sqrt{\frac{2\log(\frac{10 M D}{\delta})}{N_\T}} \sigma_r 
\end{align*}
with probability at least $1-\frac{\delta}{M}$.

\subsubsection{Proof of Lemma \ref{lemma.2}.c}
We have $W_n= \widehat{H}_n(\widehat{\Theta})-H(\widehat{\Theta})  =\widehat{\T}_n(\Theta^\star)-\widehat{\T}_n(\Theta_m)+\T(\Theta_m)-\T(\Theta^\star)$. Note that $W_n$ is identical to the term $\widehat{W}_n$ used in the proof of Lemma2.b. Thus, the proof follows by employing the same argument as presented in the proof of Lemma \ref{lemma.2}.b.

\subsection{Proof of Lemma \ref{lemma.3}}
At first, consider the derivative of the expression $N-\alpha\log (\beta N)$ with respect to $N$, which is equal to $1-\frac{\alpha}{N}$. This derivative is non-negative for $ N \geq \max \{\alpha,\;2\alpha \log(\alpha\beta)\}$. Consequently, if the inequality $N \geq \alpha\log (\beta N)$ holds for $N=\max \{\alpha,\;2\alpha \log(\alpha\beta)\}$, it also holds for $N >\max \{\alpha,\;2\alpha \log(\alpha\beta)\}$. Therefore, our objective is to establish that $N \geq \alpha\log (\beta N)$ when $N=\max \{\alpha,\;2\alpha \log(\alpha\beta)\}$. To this end, first assume that $\alpha \geq \;2\alpha \log(\alpha\beta) $. In this case, we have $N=\max \{\alpha,\;2\alpha \log(\alpha\beta)\}=\alpha$. Substituting $N=\alpha$ into $N \geq \alpha\log (\beta N)$ gives $\alpha \geq \alpha \log (\alpha \beta)$, , which is trivial since we have assumed $\alpha \geq 2\alpha \log(\alpha\beta)$.
Now, consider the scenario where $2\alpha \log(\alpha\beta) \geq \alpha$. In this case, we find that $N = \max \{\alpha,\; 2\alpha \log(\alpha\beta)\} = 2\alpha \log(\alpha\beta)$. Substituting $N = 2\alpha \log(\alpha\beta)$ into $N \geq \alpha\log(\beta N)$ leads to
\begin{equation*}
    2\alpha \log(\alpha\beta) \geq \alpha \log (2\beta\alpha\log(\alpha\beta)) \;\leftrightarrow\;  (\alpha\beta)^2 \geq 2\beta\alpha\log(\alpha\beta) \;\leftrightarrow\;   \alpha\beta \geq 2 \log(\alpha\beta).
\end{equation*} 
The last inequality holds for all $\alpha, \beta > 0$ (noting that $x \geq 2\log(x)$ for all $x > 0$).

\subsection{Proof of Lemma \ref{lemma.4}}
We prove Lemma \ref{lemma.4} via induction. The base case ($n=1$) is trivial. Now, assuming that \eqref{Eq:Sandwich} holds for iteration $n$, we will show that it also holds for iteration $n+1$. Based on the definitions of the recursions $\{Y_n\}_{n=1}^{N_e}$ and $\{Z_n\}_{n=1}^{N_e}$ in Lemma \ref{lemma.1}, we have
    \begin{equation}\label{Eq:Proof.Lemma1a. Y_n+1 and Z_n+1}
    \left[ \begin{array}{c}
      Y_{n+1} -\Theta_H \\
      Z_{n+1} -\Theta_H 
    \end{array}\right]= \left[ \begin{array}{c}
      (1-\lambda)(Y_n-\Theta_H)+\lambda (Z_n-\Theta_H)\\
      (1-\lambda) (Z_n-\Theta_H) + \lambda \left( \widehat{\T}_n (Y_{n+1})-\widehat{\T}_n(\Theta_H)+ W_n\right) 
    \end{array}\right].
\end{equation}
Since $\widehat{\T}_n$ is $\gamma-$contractive we have $-\gamma\|Y_{n+1}-\Theta_H\|_{\infty} \mathds{1} \leq-\|\widehat{\T}_n (Y_{n+1})-\widehat{\T}_n(\Theta_H)\|_{\infty} \mathds{1} \leq \widehat{\T}_n (Y_{n+1})-\widehat{\T}_n(\Theta_H) \leq \|\widehat{\T}_n (Y_{n+1})-\widehat{\T}_n(\Theta_H)\|_{\infty} \mathds{1}\leq \gamma\|Y_{n+1}-\Theta_H\|_{\infty} \mathds{1}$, which leads to
\small
\begin{equation}\label{Eq:Proof.Lemma1a.T_n}
     -\gamma\Big((1-\lambda)\|Y_n-\Theta_H\|_{\infty}+\lambda\|Z_n-\Theta_H\|_{\infty}\Big) \mathds{1} \leq \widehat{\T}_n (Y_{n+1})-\widehat{\T}_n(\Theta_H) \leq \gamma\Big((1-\lambda)\|Y_n-\Theta_H\|_{\infty}+\lambda\|Z_n-\Theta_H\|_{\infty}\Big) \mathds{1}.
\end{equation}
\normalsize
Also, from induction we have
\begin{equation}\label{Eq:Proof.Lemma1a.Induction.1}
    \left[ \begin{array}{c}
      P_{n}^y-a_{n}^y \mathds{1} \\
      P_{n}^z-a_{n}^z \mathds{1} 
    \end{array}\right] \leq  \left[ \begin{array}{c}
      Y_n -\Theta_H \\
      Z_n -\Theta_H 
    \end{array}\right] \leq \left[ \begin{array}{c}
      a_{n}^y \mathds{1}+P_{n}^y \\
      a_{n}^z \mathds{1}+P_{n}^z 
    \end{array}\right],
\end{equation}
which implies
\begin{equation}\label{Eq:Proof.Lemma1a.Induction.2}
     \left[ \begin{array}{c}
       \| Y_n -\Theta_H\|_{\infty} \\
       \| Z_n -\Theta_H\|_{\infty} 
    \end{array}\right] \leq   \left[ \begin{array}{c}
      a_{n}^y \\
      a_{n}^z 
    \end{array}\right]+ \left[ \begin{array}{c}
     \|P_{n}^y\|_{\infty} \\
      \|P_{n}^z\|_{\infty} 
    \end{array}\right].
\end{equation}
Substituting the above inequality into \eqref{Eq:Proof.Lemma1a.T_n} gives
\begin{align}\label{Eq:Proof.Lemma1a.T_n.2}
     -\gamma\Big((1-\lambda)(a_n^y+\|P_n^y\|_{\infty})+\lambda(a_n^z+\|P_n^z\|_{\infty})\Big) \mathds{1} &\leq \widehat{\T}_n (Y_{n+1})-\widehat{\T}_n(\Theta^\star)\nonumber\\   
  &\leq \gamma\ \Big((1-\lambda)(a_n^y+\|P_n^y\|_{\infty})+\lambda(a_n^z+\|P_n^z\|_{\infty})\Big) \mathds{1}.
\end{align}
By substituting \eqref{Eq:Proof.Lemma1a.Induction.1}, \eqref{Eq:Proof.Lemma1a.Induction.2} and \eqref{Eq:Proof.Lemma1a.T_n.2} into \eqref{Eq:Proof.Lemma1a. Y_n+1 and Z_n+1} we find that
\small
\begin{align*}
   &\left[ \begin{array}{c}
      \underbrace{ -\Big((1-\lambda) a_n^y+\lambda a_n^z\Big)}_{-a_{n+1}^y}\mathds{1}+\underbrace{ (1-\lambda)P_n^y+\lambda P_n^z }_{P_{n+1}^y} \\      
     \underbrace{-\Big((1-\lambda+\lambda^2\gamma) a_n^z + \gamma(1-\lambda)\lambda a_n^y+\gamma\lambda\big((1-\lambda)\|P_n^y\|_{\infty}+\lambda \|P_n^z\|_{\infty}\big)\Big)}_{-a_{n+1}^z}\mathds{1}+\underbrace{(1-\lambda)P_n^z+\lambda W_n}_{P_{n+1}^z}
    \end{array}\right]  \\
    &\qquad\qquad\qquad\qquad\qquad\qquad\qquad\qquad\leq \left[ \begin{array}{c}
      Y_{n+1} -\Theta_H \\
      Z_{n+1} -\Theta_H 
    \end{array}\right]\leq \\
    &\left[ \begin{array}{c}
      \underbrace{ \Big((1-\lambda) a_n^y+\lambda a_n^z\Big)}_{a_{n+1}^y} \mathds{1} +\underbrace{ (1-\lambda)P_n^y+\lambda P_n^z }_{P_{n+1}^y} \\      
     \underbrace{\Big((1-\lambda+\lambda^2\gamma) a_n^z + \gamma(1-\lambda)\lambda a_n^y+\gamma\lambda\big((1-\lambda)\|P_n^y\|_{\infty}+\lambda \|P_n^z\|_{\infty}\big)\Big)}_{a_{n+1}^z}\mathds{1}+\underbrace{(1-\lambda)P_n^z+\lambda W_n}_{P_{n+1}^z}
    \end{array}\right],
\end{align*}
\normalsize
which completes the proof.

\subsection{Proof of Lemma \ref{lemma.5}}
According to Lemma \ref{lemma.4}, we have
\begin{equation*}
    \left[ \begin{array}{c}
      a_{n+1}^y \\
      a_{n+1}^z 
    \end{array}\right]=H  \left[ \begin{array}{c}
      a_{n}^y \\
      a_{n}^z 
    \end{array}\right]+F \left[ \begin{array}{c}
      \|P_n^y \|_{\infty}\\
      \|P_n^z\|_{\infty} 
    \end{array}\right].
\end{equation*}
Using the above inequality, we find that 
\begin{align*}
    \sum_{n=1}^{N_e} \left[ \begin{array}{c}
      a_{n}^y \\
      a_{n}^z 
    \end{array}\right] = \left[ \begin{array}{c}
      a_{1}^y \\
      a_{1}^z 
    \end{array}\right]+\sum_{n=1}^{N_e-1} \left[ \begin{array}{c}
      a_{n+1}^y \\
      a_{n+1}^z 
    \end{array}\right]= & \left[ \begin{array}{c}
      a_{1}^y \\
      a_{1}^z 
    \end{array}\right] + H \sum_{n=1}^{N_e} \left[ \begin{array}{c}
      a_{n}^y \\
      a_{n}^z 
    \end{array}\right]+F\sum_{n=1}^{N_e} \left[ \begin{array}{c}
      \|P_n^y \|_{\infty} \\
      \|P_n^z\|_{\infty} 
    \end{array}\right]\\
    &- H \left[ \begin{array}{c}
      a_{N_e}^y \\
      a_{N_e}^z 
    \end{array}\right]-F\left[ \begin{array}{c}
      \|P_{N_e}^y \|_{\infty} \\
      \|P_{N_e}^z\|_{\infty} 
    \end{array}\right],
    \end{align*}
     which implies
     \begin{equation}\label{Eq:Lemma1b.b}
     \sum_{n=1}^{N_e} \left[ \begin{array}{c}
      a_{n}^y \\
      a_{n}^z 
    \end{array}\right]=(1-H)^{-1} \left( \left[ \begin{array}{c}
      a_{1}^y \\
      a_{1}^z 
    \end{array}\right]+\sum_{n=1}^{N_e} \left[ \begin{array}{c}
      \|P_n^y \|_{\infty} \\
      \|P_n^z\|_{\infty} 
    \end{array}\right]-H \left[ \begin{array}{c}
      a_{N_e}^y \\
      a_{N_e}^z 
    \end{array}\right]-F\left[ \begin{array}{c}
      \|P_{N_e}^y \|_{\infty} \\
      \|P_{N_e}^z\|_{\infty} 
    \end{array}\right]  \right).
   \end{equation}
Note that all the entries of the state matrix $H$ and the input matrix $F$ are between zero and one. Moreover, we have
\begin{equation*}
(1-H)^{-1}= \frac{1}{\operatorname{det}(1-H)}\left[\begin{array}{cc}
    1-H_{22} & H_{12}  \\
     H_{21} & 1-H_{11}
\end{array} \right],
\end{equation*}
where $\operatorname{det}(1-H)=\lambda^2 (1-\gamma) > 0$. Therefore, $(1-H)^{-1}$ Also has positive entries. Hence, 
\begin{equation*}
  (1-H)^{-1}\left( H \left[ \begin{array}{c}
      a_{N_e}^y \\
      a_{N_e}^z 
    \end{array}\right]+ F \left[ \begin{array}{c}
      \|P_{N_e}^y \|_{\infty} \\
      \|P_{N_e}^z\|_{\infty} 
    \end{array}\right] \right) \geq 0.
\end{equation*}
By adding the above expression to the right-hand side of \eqref{Eq:Lemma1b.b}, we obtain 
   \begin{align*}
    \sum_{n=1}^{N_e} \left[ \begin{array}{c}
      a_{n}^y \\
      a_{n}^z 
    \end{array}\right] \leq (1-H)^{-1} \left( \left[ \begin{array}{c}
      a_{1}^y \\
      a_{1}^z 
    \end{array}\right] + \sum_{n=1}^{N_e} \left[ \begin{array}{c}
      \|P_n^y \|_{\infty} \\
      \|P_n^z\|_{\infty}
      \end{array}\right] \right),
      \end{align*} 
as claimed.
\end{appendices}

\printbibliography

\end{document}